\documentclass{article} 
\PassOptionsToPackage{table}{xcolor}
\usepackage[final]{colm2025_conference}

\usepackage{microtype}
\usepackage{hyperref}
\usepackage{url}
\usepackage{booktabs}
\usepackage{graphicx}
\usepackage{amsmath}
\usepackage{amssymb}
\usepackage{tabularx}
\usepackage{pifont}
\usepackage{array}
\usepackage{multirow}
\usepackage{siunitx}
\usepackage{listings}
\usepackage{subcaption}
\usepackage{lineno}
\usepackage{makecell}
\usepackage{pifont}
\usepackage{xspace}
\usepackage{lipsum}
\usepackage{wrapfig}
\usepackage[table]{xcolor}

\definecolor{darkblue}{rgb}{0, 0, 0.5}
\hypersetup{colorlinks=true, citecolor=darkblue, linkcolor=darkblue, urlcolor=darkblue}

\newcommand{\aspace}{\hspace{1em}}

\title{\includegraphics[height=1.2em]{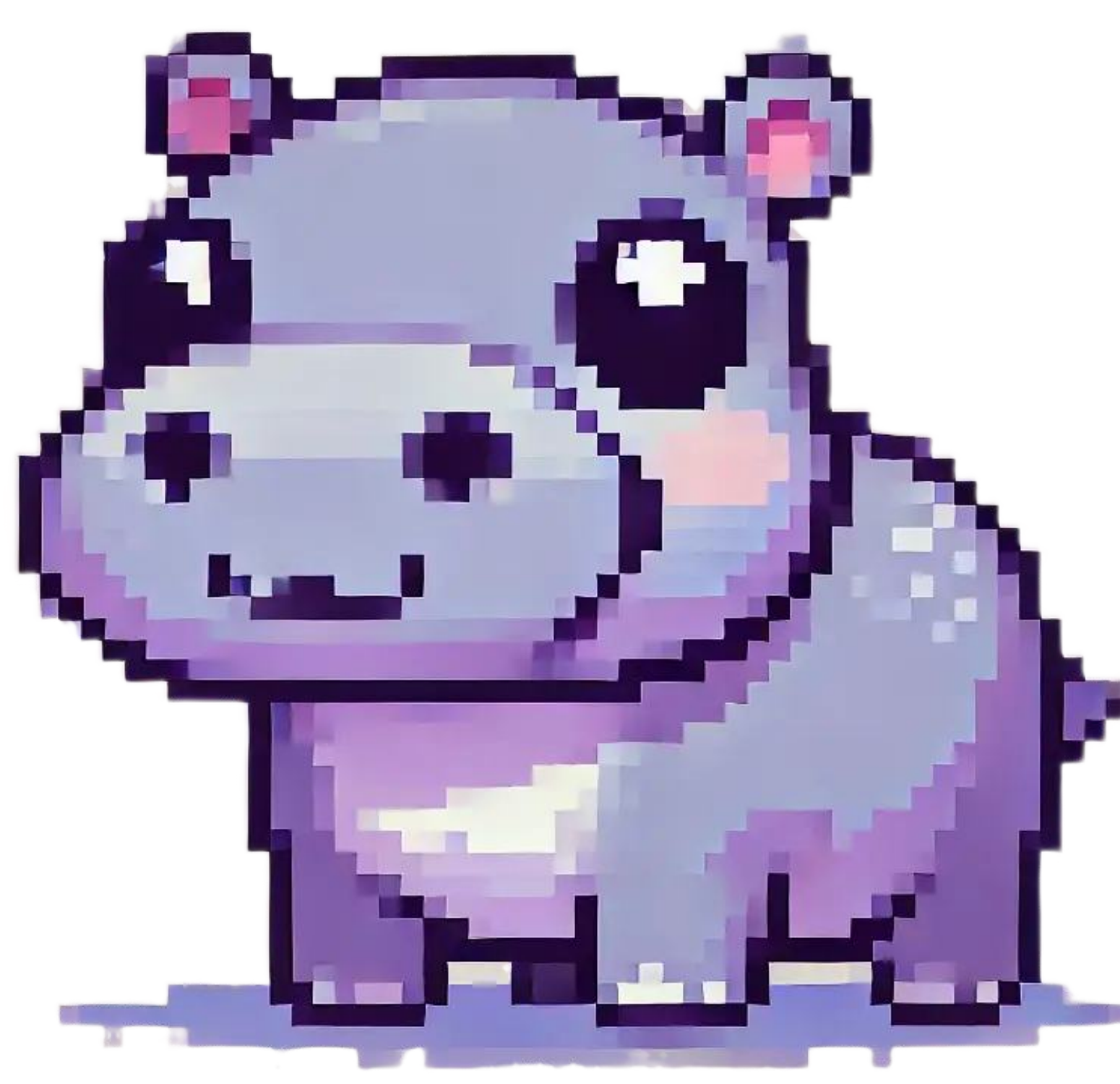} \dataset : Simulating Watch Histories with Large Language Models for Personalized Video Highlighting}

\author{Jeongeun Lee \aspace Youngjae Yu \aspace Dongha Lee\thanks{Corresponding Author} \\
Yonsei University\\
\texttt{\{ljeadec31, yjy, donalee\}@yonsei.ac.kr}
}

\newcommand{\task}{{personalized video highlighting}\xspace}
\newcommand{\dataset}{\textsc{HiPpo-Video}~}
\newcommand{\datasets}{\textsc{HiPpo-Video}'s\xspace}
\newcommand{\dataseth}{\textsc{HiPpo-Video}\textsuperscript{+}}
\newcommand{\proposed}{{HiPHer}\xspace}

\newcommand{\cmark}{\checkmark} 
\newcommand{\xmark}{\ding{55}}   
\newcommand{\smallsection}[1]{{\vspace{0.05in} \noindent 
\bf {#1.\hspace{5pt}}}}

\begin{document}

\ifcolmsubmission
\linenumbers
\fi

\maketitle

\begin{abstract}
The exponential growth of video content has made personalized video highlighting an essential task, as user preferences are highly variable and complex. 
Existing video datasets, however, often lack personalization, relying on isolated videos or simple text queries that fail to capture the intricacies of user behavior. 
In this work, we introduce \dataset, a novel dataset for personalized video highlighting, created using an LLM-based user simulator to generate realistic watch histories reflecting diverse user preferences.
The dataset includes 2,040 (watch history, saliency score) pairs, covering 20,400 videos across 170 semantic categories.
To validate our dataset, we propose \proposed, a method that leverages these personalized watch histories to predict preference-conditioned segment-wise saliency scores. 
Through extensive experiments, we demonstrate that our method outperforms existing generic and query-based approaches, showcasing its potential for highly user-centric video highlighting in real-world scenarios.

\begin{center}
\begin{tabular}{cll}
\raisebox{-1.5pt}{\includegraphics[height=1.05em]{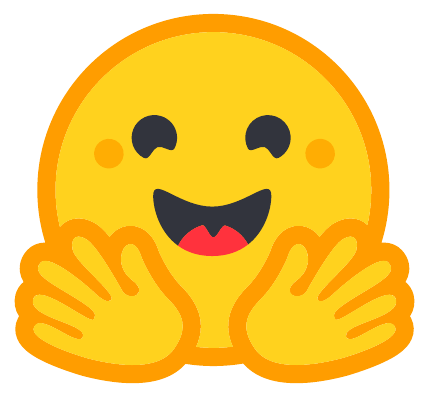}} & \textbf{Dataset:} &  \href{https://huggingface.co/datasets/jeongeunnn/HIPPO-video} {\path{huggingface.co/datasets/jeongeunnn/HIPPO-video}}\\
\raisebox{-1.5pt}{\includegraphics[height=1.05em]{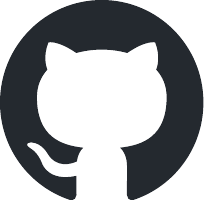}} & \textbf{Code:} & \href{https://github.com/jeongeunnn-e/HIPPO-Video}{\path{github.com/jeongeunnn-e/HIPPO-Video}}\\
\end{tabular}
\end{center}

\end{abstract}

\section{Introduction}
As the scale and diversity of video content rapidly grow in the real world, it becomes increasingly important for users to digest long-form videos efficiently within limited time and resources~\citep{huang2020movienet, apostolidis2021video, argaw2024towards}. 
In this context, various research tasks have emerged to generate shorter, more consumable versions of videos—such as video summarization~\citep{park2020sumgraph, xu2024mh}, highlight detection, and moment retrieval~\citep{lin2023univtg, sun2024tr, xiao2024bridging, xu2024mh}.

However, these tasks often overlook the importance of personalization in the real world where \textit{important} moments vary significantly among users.
Tailoring to individual interests can better meet the demand for user-centric content delivery than a one-size-fits-all approach.
While some prior works in query-focused video summarization~\citep{vasudevan2017query, xiao2020query, xiao2020convolutional} and moment retrieval~\citep{liu2018attentive, zeng2022moment} have explored aspects of personalization, they typically reduce user preferences to a single phrase or feature, oversimplifying the complexity of human interest.
In reality, human preferences are multifaceted, evolving over time and across different types of content.
To address this, we propose leveraging watch history as a richer source of user preference modeling.
We contend that analyzing users’ sequential viewing behavior through their watch histories can uncover underlying preferences, leading to more effective and tailored video experiences.

\begin{figure*}[t]
    \centering    
    \includegraphics[width=\linewidth]{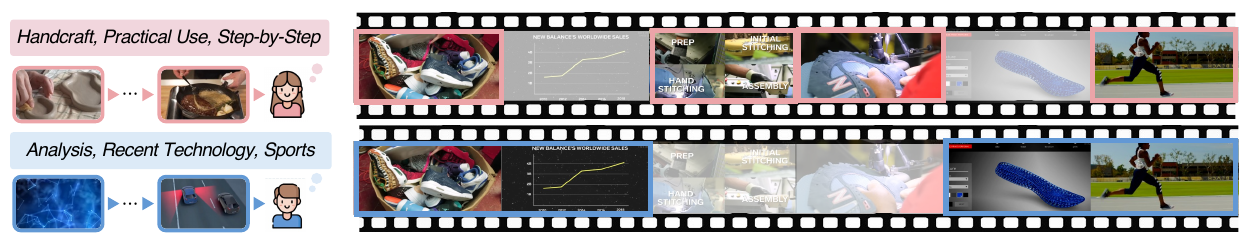} 
    \caption{A video can produce varying highlights based on user interests, showing how watch history reflects implicit feedback and helps tailor highlights to individual preferences.}
    \label{fig:intro}
\end{figure*}

In this work, we introduce \textbf{\task}, a novel task that leverages a user’s watch history within a single session to tailor video highlights to the user's preferences.
Inspired by how recommender systems effectively capture user interests through implicit feedback, such as interaction history~\citep{rendle2009bpr, kang2018self}, our task aims to dynamically select and present highlight segments aligned with the user’s real-time viewing behavior and preferences during the session.
For instance, as shown in Figure~\ref{fig:intro}, the same video may yield distinct highlights depending on the user’s focus inferred from their watch history, emphasizing different aspects of content.

For this task, we introduce \dataset: \textbf{\underline{Hi}}ghlights Based on \textbf{\underline{P}}references for \textbf{\underline{P}}ersonalized Vide\textbf{\underline{O}} Clipping, a large-scale dataset containing user watch histories and corresponding personalized saliency scores, generated by simulating real-world user behavior on video platforms. 
Existing video datasets~\citep{gygli2014creating, song2015tvsum, sharghi2016query} are often limited in scale due to resource-intensive nature of manual annotation, while collecting actual users' watch histories raises privacy concerns.
To address these challenges, we leverage large language models (LLMs) to simulate user interactions, enabling scalable data generation without compromising user privacy.
\dataset consists of 2,040 (watch history, saliency score) pairs, where each watch history comprises 10 videos, thereby totaling 20,400 videos, across 170 semantic initial preference seeds.
 
Through experiments, we validate our task and dataset using a simple baseline, \textbf{\underline{Hi}}story-Driven \textbf{\underline{P}}reference-Aware Video \textbf{\underline{H}}ighlight\textbf{\underline{er}}, named \textbf{\underline{\proposed}}, which leverages user preferences derived from watch history as preference context. 
\proposed outperforms existing methods by incorporating personalized preference embeddings from watch histories, while generic methods often fail to align with individual user interests, and query-focused methods struggle to capture the complexity of preferences with short queries.
These results underscore the importance of incorporating detailed user histories to enhance user-specific video highlighting, demonstrating the effectiveness of history-driven preference modeling.

\section{Related Work}

\smallsection{Tasks and Datasets}
Highlight detection identifies the most engaging or significant moments within a video by assigning importance scores to segments. 
Existing datasets~\citep{sun2014ranking, song2016click, gygli2016video2gif, sul2023mr} provide query-unrelated highlight clips.
Moment retrieval locates specific time spans within videos that match a given natural language query, using datasets~\citep{lei2020tvr, gao2017tall, lei2021detecting, zala2023hierarchical} that pair queries with annotated moments.
Video summarization provides condensed versions of videos, preserving essential narrative or informational content.  Traditional datasets~\citep{gygli2014creating, song2015tvsum, sharghi2016query} rely heavily on human annotations, limiting their scalability. Recent work has thus explored automated summarization using large language models (LLMs)~\citep{argaw2024scaling, hua2024v2xum}.
Notably, several datasets are utilized across multiple tasks~\citep{song2015tvsum, sul2023mr, lei2021detecting}. Detailed comparisons of these datasets are provided in Table~\ref{tbl:comparison}.

\smallsection{Methods}
Prior work on highlight detection has mainly explored ranking-based methods that assign scores to video segments to identify highlights~\citep{sun2014ranking, gygli2016video2gif, yao2016highlight, rochan2020adaptive}.
For moment retrieval, research has centered on cross-modal alignment techniques to bridge textual queries and visual content~\citep{lu2019debug, yuan2019semantic, zhang2020span, lei2020tvr}.
More recently, inspired by the success of DETR~\citep{carion2020end}, DETR-based methods have been proposed to jointly tackle both moment retrieval and highlight detection in a unified framework~\citep{lei2021detecting, moon2023query, liu2022umt}.
On the other hand, video summarization selects key moments to represent overall content~\citep{ji2019video, argaw2024scaling}. When a query is given~\citep{sharghi2016query, sharghi2017query, narasimhan2021clip}, it becomes query-focused summarization, similar to moment retrieval in aligning video content with textual input.

\begin{table*}[tbp]
    \centering
    \small
    \resizebox{0.99\linewidth}{!}{
    \begin{tabular}{l|cccccc}
        \toprule
        \multirow{2.5}{*}{Dataset} 
        & \multicolumn{2}{c}{Statistics}
        & \multirow{2.5}{*}{\makecell{Supported\\Tasks}} 
        & \multicolumn{2}{c}{Single Instance} 
        & \multirow{2}{*}{Anno.} \\ 
        \cmidrule(lr){2-3} \cmidrule(lr){5-6}  
        & \#Videos & Avg Len(m) & & Query & \#Videos \\ 
        \midrule
        YouTubeHighlights~\citep{sun2014ranking} & 600 & 2.4 & MR, HD & \xmark & 1 & M \\
        SumMe~\citep{gygli2014creating} & 25 & 2.4 & VS & \xmark & 1 & M\\
        TVSum~\citep{song2015tvsum} & 50 & 3.9 & VS & \xmark & 1 & M \\
        QFVS~\citep{sharghi2016query} & 4 & 240 & VS, MR & \cmark & 1 & M \\
        Charades-STA~\citep{gao2017tall} & 6,700 & 0.5 & MR & \cmark & 1 & M \\
        TVR~\citep{lei2020tvr} & 21,800 & 1.3 & MR & \cmark & 1 & M \\
        QVHighlights~\citep{lei2021detecting} & 10,200 & 2.5 & MR, VS & \cmark & 1 & M \\
        Mr.HiSum~\citep{sul2023mr} & 31,892 & 3.4 & VS, HD & \xmark & 1 & M \\
        Shot2Story20K~\citep{han2023shot2story20k} & 20,023 & 0.3 & VS & \xmark & 1 & M+S \\
        Instruct-V2Xum~\citep{hua2024v2xum} & 30,000 & 3.1 & VS & \xmark & 1 & M+S \\
        LfVS~\citep{argaw2024scaling} & 1,200 & 12.2 & VS & \xmark & 1 & M+S \\
        \midrule
        \includegraphics[height=1em]{FIG/hippo.pdf} \textbf{\dataset} & 2,040(20,400) & 13.9 & VS, MR, HD, PV & \cmark & 10 & M+S\\ 
        \bottomrule
    \end{tabular}
    }
   \caption{Benchmark dataset comparison across tasks: \textbf{VS} (Video Summarization), \textbf{MR} (Moment Retrieval), \textbf{HD} (Highlight Detection), and \textbf{PV} (Personalized Video Highlighting). \textbf{M} denotes manually curated datasets, while \textbf{S} refers to those synthesized by models.}
    \label{tbl:comparison}
\end{table*}

\section{\dataset}
We introduce \dataset, a large-scale dataset designed for personalized video highlighting. 
The dataset consists of (1) user watch history sequences and (2) 10-point saliency scoring annotations for target video. 
Each sequence consists of 10 videos and the dataset includes 2,040 sequences, resulting in a total of 20,400 videos {across a variety of categories.}

\subsection{Simulation}
\label{subsec:simulation}

\begin{figure*}[t]
    \centering    
    \includegraphics[width=\linewidth]{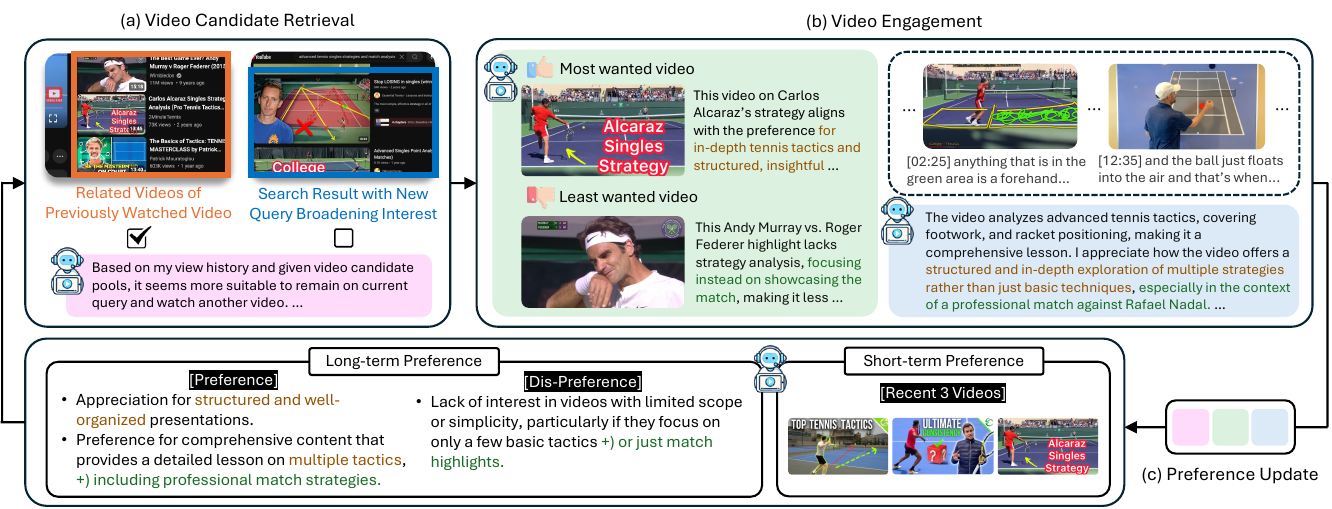} 
    \caption{The overall process of our LLM-based user simulation to collect video watch histories on a video platform operates iteratively as follows: (1) retrieving video candidates either from related videos or through a new query, (2) engaging with videos, including selection and viewing, and (3) updating long-term preferences based on the simulator's key responses obtained during video retrieval and engagement.}
    \label{fig:overview}
\end{figure*}

Collecting real user watch histories from video platforms presents significant challenges, including privacy concerns and resource constraints. 
To address these limitations, we employ LLM-based user simulator\footnote{Hereafter, we refer to it as ``the simulator'’ for brevity.} to generate realistic, large-scale video watch history sequences.
Figure~\ref{fig:overview} provides an overview of the watch history simulation process, and detailed prompts are included in Appendix~\ref{apdx:prompts}.

Starting from an initial profile seed, the simulator operates iteratively, dynamically updating user preferences as it \textit{watches} videos.
Specifically, the process consists of three steps : (1) video candidate retrieval, (2) video engagement, and (3) preference update. 
This iterative framework enables the simulator to capture the evolving nature of real user preferences, effectively modeling the complexity and diversity of real-world video consumption.\footnote{In our simulation, we set the number of video candidates per turn to $l=8$ and fix the number of watched videos in history to $m=10$.}

\smallsection{Initialization}
To support diversity in simulating user behavior, we initialize simulators with carefully designed variables representing user interests.
These variables follow the categorization from \citet{qiu2024mmsum}, comprising 170 topic and sub-topic pairs adapted from existing video datasets and popular Wikipedia topics~\citep{zhou2018towards, miech2019howto100m}, capturing the breadth of content on YouTube.
Additionally, we introduce a sentiment-based variable (\textit{intent}) to model user motivations and viewing preferences.
By integrating topic categorization and intent-informed preferences, we construct 2,040 profiles as initial seeds for personalized watch history simulation, contributing to adaptability across diverse users and video content. 
Details on initialization variables are provided in the Appendix~\ref{apdx:user_init_profile}.


\smallsection{Video Candidate Retrieval}
The simulation begins with retrieving a set of video candidates, denoted as $\mathcal{C} = \{C_1, C_2, \dots, C_l\}$, by crawling YouTube {in real time}. 
In real-world viewing sessions, users typically either continue exploring within a topic or shift (or expand) to new topics. 
To model this behavior, the simulator is given two options: (1) exploring related videos or (2) generating a new search query. 
Specifically, at the $i$-th turn (i.e., $i$-th video selection), this decision is informed by previously watched videos, $\mathcal{H}_{i-1} = \{ H_1, H_2, \dots, H_{i-1}\}$, along with the user’s {current} preference, $p_{i-1}$.
This approach enables the simulator to simulate natural browsing patterns, balancing topic continuity and exploration.

\smallsection{Video Engagement}
Once the candidate pool is retrieved, the simulator selects a video to watch and engages with its content. 
First, the selection process accounts for two types of user preferences: \textit{short-term} and \textit{long-term}.
Short-term preferences are based on the metadata\footnote{Metadata includes information identical to that displayed on YouTube, such as the title, channel, thumbnail, and view count. Detailed examples are provided in the Appendix~\ref{apdx:youtube_crawling}.} of 3 most recently watched videos, while long-term preferences, denoted as $p_{i-1},$ are expressed as explicit likes and dislikes in natural language, providing an accumulated profile of the user’s overall interests over the video sequence $\mathcal{H}_{i-1}$.
To refine preference modeling, the simulator selects both the most wanted video and the least wanted one.
This contrastive approach enhances user modeling by building a more fine-grained representation of preference, balancing both likes and dislikes.
Additionally, the simulator provides reasoning for its selections (green box in Figure~\ref{fig:overview}), reinforcing the decision-making process.

Once the most preferred video $C\in\mathcal{C}$ is determined, the simulator proceeds to \textit{watch} it. 
The video is segmented into $C = \{s_1, s_2, \dots, s_n\}$ using scene change detection,\footnote{\url{https://github.com/Breakthrough/PySceneDetect}} ensuring each segment forms a coherent unit of content.
Each segment is represented as $s_k = (v_k, t_k)$, where $v_k$ is visual description and $t_k$ is the corresponding transcript.
To facilitate comprehensive video understanding for LLM~\citep{wang2024videoagent}, the representative frame $f_k$ in segment $s_k$ is converted into textual description $v_k$ via frame captioning with \citet{liu2024visual}. 
Using this multimodal input, the simulator generates a \textit{review}, which includes a concise summary and tailored opinion on the video, aligned with the evolving preferences $p_{i-1}$ (blue box in Figure~\ref{fig:overview}).
This entire process—\textit{selecting} and \textit{watching} a video—replicates the human process of interacting with content, guided by both recent interactions and long-term interests.

\smallsection{Preference Update}
After completing video engagement, the simulator updates its preference state from $p_{i-1}$ to $p_i$.
During the engagement, the simulator generates three key responses based on preference reasoning: (1) the rationale for selecting the most preferred video, (2) the rationale for selecting the least preferred video, and (3) a review of the watched video.
These are then used to refine long-term preferences, dynamically adjusting based on recent interactions.
In Figure~\ref{fig:overview}, additional details inferred from the simulator’s key responses (highlighted in brown and green) are incorporated into the long-term preference.

\subsection{Saliency Score Annotation}
\label{subsec:score_anno}
After simulation, the last video in each watch history is set as the target video for saliency annotation. 
Similar to the video engagement process, the video is segmented by detecting scene changes.
The simulator then assigns relevance scores ranging from 1 to 10 to each segment.
These scores are determined based on two primary sources of information: (1) final long-term preferences, consolidated  over the course of watching the videos, and (2) personal reviews generated each time after watching the video.
The review-driven preferences offer video-specific signals, while the long-term preferences reflect broader interests across entire session.
By integrating these two preference layers, the simulator establishes session-based user inclinations, enabling the segment scoring that aligns with inferred interests.

\subsection{Human Verification}
\label{ref:human_verification}
\smallsection{Validation of Watch History Simulation Process}
To evaluate the reliability of our watch history simulation, we employ Amazon Mechanical Turk (MTurk) annotators to assess two key aspects of the framework: (1) query generation and (2) video selection. 
Annotators are given the same preference information as the LLM-based user simulator, including previously watched videos and long-term user preferences.
For query generation, annotators assess whether queries written by the simulator are plausible for next steps. Results show that 97.56\% of queries are reasonable, with 85\% inter-annotator agreement.
For video selection, annotators are given a set of candidate videos, identical to the simulator’s pool, and asked to select the one that best aligns with the provided preferences.  
The simulator’s choices match human selections in {71.42\%} of cases, suggesting that it effectively mirrors real user behavior.
More details on the evaluation process can be found in Appendix~\ref{apdx:human_verf}.

\begin{wraptable}{r}{0.43\textwidth}
    \centering
    \small
    \vspace{-0.7em}
    \resizebox{0.99\linewidth}{!}{
    \begin{tabular}{c|ccc|c}
        \toprule
        Agreement & A & U & D & Percentage \\ 
        \midrule
        \multirow{3}{*}{Agree} & 3 & 0 & 0 & \textbf{64.10\%} \\
         & 2 & 1 & 0 & 15.38\% \\
         & 2 & 0 & 1 & 17.95\% \\
         \midrule
        Neutral & 1 & 2 & 0 & 2.56\% \\
        \bottomrule
    \end{tabular}
    }
    \caption{User agreement results}
    \label{tbl:user_study}
    \vspace{-1em}
\end{wraptable}
\textbf{Validation of Saliency Annotations. }
To validate the saliency annotations generated by the simulator, we conduct a user study adapting the methodology from \cite{sul2023mr}.
MTurk annotators are given a video, a user preference, and the clip assigned the highest saliency score (or multiple pairs if there are tied clips). 
For each pair, the annotators determine whether highlighted clip aligns with the given preference by selecting one of three options: Agree (A), Unclear (U), or Disagree (D). In Table~\ref{tbl:user_study}, the A, U, and D columns represent the number of annotators who selected each option. 
The results show that nearly 98\% of pairs are deemed reasonable by majority agreement, confirming that the saliency scores accurately capture personalized preferences.

\smallsection{Validation of Simulated Watch Histories}
To further verify the realism of our simulated watch histories, we conduct complementary evaluations using 40 real user histories (10 videos each), collected with informed consent.\footnote{We refer to this dataset as \dataseth. Human annotators recorded their watch histories and annotated the final video at the segment level, following the procedure in Section~\ref{subsec:score_anno}. This dataset is also used for evaluating baselines in Section~\ref{subsec:exp_dataseth}.}
First, following recent LLM-as-a-judge protocols~\citep{chiang2024chatbot, mitchell2023detectgpt, luo2025videoautoarena}, we task GPT-4~\citep{achiam2023gpt} with binary classification to distinguish simulated from real histories initialized with the same profile seed. GPT-4 achieves only 40\% accuracy, below the 50\% random baseline, indicating that the simulated histories are often indistinguishable from real ones.
Second, we apply Fast-DetectGPT~\citep{bao2023fast} in a Hit@1 setting, where the model must identify one simulated history from a set of nine real ones. It achieves a Hit@1 score of 0.350, indicating substantial confusion and further supporting the similarity between simulated and real histories.
Together, these results strongly support the validity of our simulation framework as a reliable proxy for real-world user watch histories.

\subsection{Dataset Analyses}

We analyze key aspects of our dataset, including its overall characteristics and diversity, as shown in Figure~\ref{fig:dataset_anal}, with detailed analysis settings provided in Appendix~\ref{apdx:human_verf}.

\smallsection{Overall Statistics}
\dataset is thoroughly curated from real-time crawling, with all videos in the watch history sequences spanning from 2008 to 2024.
Of these, 57.16\% were published after 2023, ensuring the dataset remains up-to-date.
Video durations range from 30 seconds to 119 minutes, with an average length of 13.9 minutes, reflecting typical video consumption.
For annotation, target videos are divided into an average of 56.91 segments.

\smallsection{Intra-history Video Diversity}
We analyze the \textit{exploration ratio} within a video watch history to measure how actively the simulator broadens its interests. 
When the simulator chooses to watch related videos or repeats a similar query to previous ones, this is considered as non-exploration. 
In contrast, when the simulator generates a distinctly new query, we define this as \textit{topic drift}, signifying an expansion of interest. 
As shown in Figure~\ref{fig:explorate}, the exploration rate generally ranges from 0.2 to 0.6, indicating a wide spectrum of behavioral patterns among simulators—some maintaining consistent, focused interests, while others frequently shift topics and explore new content areas.

\smallsection{Inter-history Video Diversity}
To assess the diversity of user preferences captured through simulation, we visualize the embedding space of video watch histories using t-SNE, with embeddings generated by CLIP~\citep{radford2021learning}.
As shown in Figure~\ref{fig:history_tsne}, the embeddings do not form tight clusters or align strictly with their initial topics. 
This indicates that the simulated watch histories encompass a wide range of user preferences, even when originating from predefined topics (as explained in Section~\ref{subsec:simulation} \textit{Initialization}).

\smallsection{Saliency Score Distribution}
Figure~\ref{fig:score_distb} shows the distribution of saliency scores with kernel density estimation (KDE). 
The mean saliency scores (left) represent the average segment scores per video, with most falling between 4 and 6, indicating moderate relevance. 
To assess fluctuations, we measure the standard deviation (right), which typically ranges from 1.5 to 2, suggesting moderate variability. 
Higher deviations (greater than 3) indicate substantial fluctuations, likely due to dynamic visual changes or frequent scene transitions.

\begin{figure*}[t]
    \centering
    \begin{subfigure}[b]{0.24\textwidth}
        \centering
        \includegraphics[width=\textwidth]{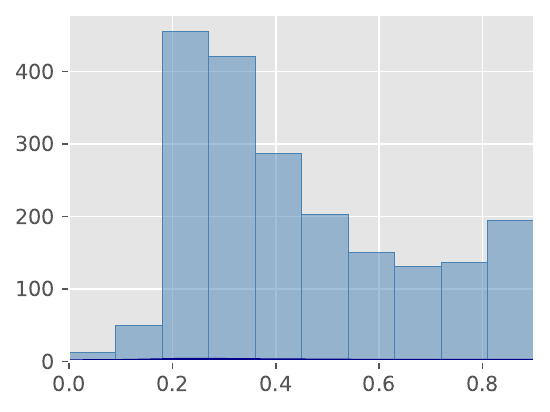}
        \caption{Exploration Ratio}
        \label{fig:explorate}
    \end{subfigure}
    \hfill             
    \begin{subfigure}[b]{0.24\textwidth}
        \centering
        \includegraphics[width=\textwidth]{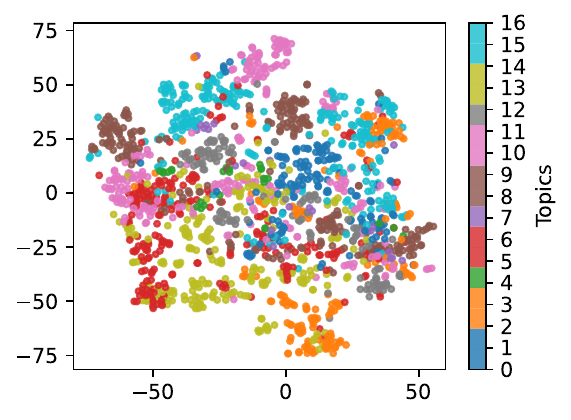}
        \caption{History Embedding}
        \label{fig:history_tsne}
    \end{subfigure}
    \hfill                                                             
    \begin{subfigure}[b]{0.48\textwidth}
        \centering
        \includegraphics[width=\textwidth]{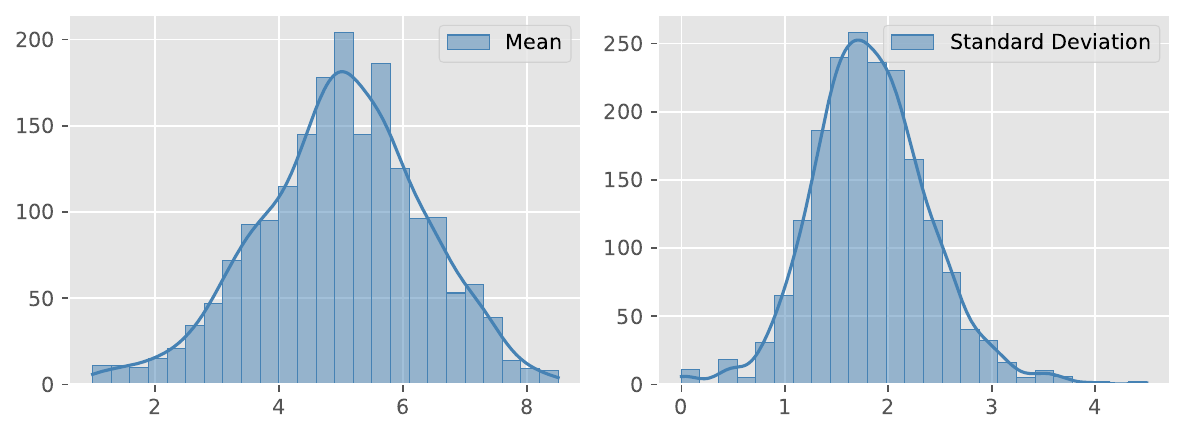}
        \caption{Saliency Score Distribution}
        \label{fig:score_distb}
    \end{subfigure}
    \caption{Dataset analysis results. (a–b) Exploration patterns and watch history embeddings visualized via t-SNE. (c) Distribution of saliency score means and standard deviations.}
    \label{fig:dataset_anal}
\end{figure*}

\section{\proposed : History-driven Preference-aware Video Highlighter}


We propose \proposed, which generates personalized segment-wise saliency scores by modeling user preferences from their watch history. 
Given a video $V$ uniformly divided into $n$ segment, and a watch history consisting of $m$ videos, $\mathcal{H} = \{ H_1, H_2, \dots, H_m \}$, the objective is to predict saliency scores $Y = \{ y_1, y_2, \dots, y_n \}$, where each $y_k$ quantifies the relevance of the $k$-th segment to the user preferences.
{\proposed derives a global preference embedding from the watch history to guide segment representations via cross-attention, producing relevance scores optimized with a contrastive loss to prioritize segments aligned with user interests.}


\smallsection{Input Representations}
\label{subsec:video-encoding}
For each of the $n$ segments of $V$, we denote the representative frames as $\{ f_1, f_2, \dots, f_n \}$, and the corresponding transcripts as $\{ t_1, t_2, \dots, t_n \}$.
The transcripts are generated from audio using Audio Speech Recognition (ASR), as ASR has shown to enhance visual recognition tasks~\citep{li2020hero}.
We employ the pre-trained CLIP image encoder(ViT-B/32)~\citep{radford2021learning} to generate visual features $\{ s^{f}_1, s^{f}_2, \dots, s^{f}_n \}$ for each frame.
Similarly, we use the CLIP text encoder to convert the transcripts into textual features $\{ s^t_1, s^t_2, \dots, s^t_n \}$.
Since visual and textual features may carry distinct semantic information,  we concatenate them for each segment $s_k = s^f_k \oplus s^t_k$, where $s^f_k$ and $s^t_k$ represent the visual and textual features, respectively, rather than directly fusing them~\citep{kamath2021mdetr}.


\begin{figure*}[t]
    \centering    
    \includegraphics[width=\linewidth]{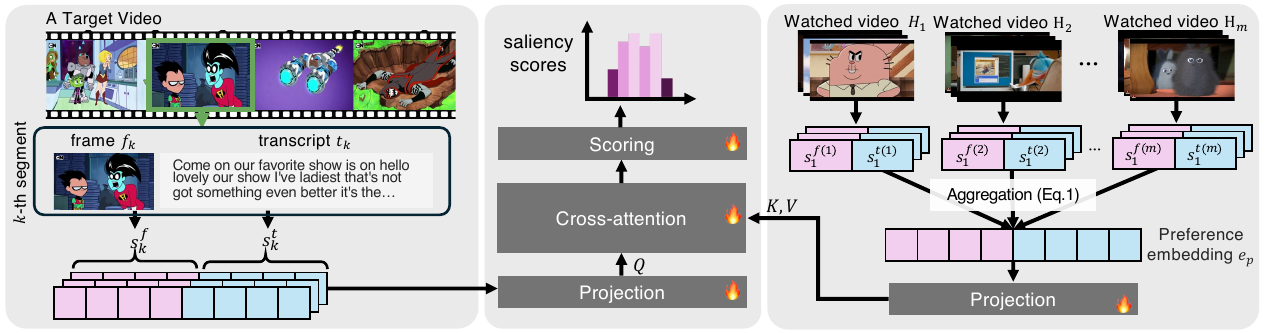}
    \caption{The architecture of \proposed consists of two modules: (1) a preference modeling module that generates a preference embedding from watched videos, and (2) a scoring module that assigns a preference score to each segment in the target video.}
    \label{fig:sumamrizer}
\end{figure*}

\smallsection{Preference Modeling from Watch History}
\label{subsubsec:prefextraction}
\proposed first constructs a preference embedding, denoted as $e_p$, to encapsulate preferences inferred from a sequence of previously watched videos $\mathcal{H}$.
Each video $H_i$ in the watch history is encoded by aggregating its segment features $\{s_1^{(i)}, s_2^{(i)}, \dots, s_n^{(i)}\}$ using $\mathcal{A}gg_s$, where each segment is encoded similarly to the target video representations.
This results in a single embedding $h^{(i)}$, which serves as a compact representation of the entire video.
The video embeddings, $\{h^{(1)}, h^{(2)}, \dots, h^{(m)}\}$, are then aggregated into a global preference representation using $\mathcal{A}gg_h$, as follows:
\begin{equation}
\small
e_p = \mathcal{A}gg_h\left( \{ h^{(i)} : h^{(i)} = \mathcal{A}gg_s(s_1^{(i)}, \dots, s_n^{(i)}) \}_{i=1}^m \right)
\end{equation}
In this work, we use mean pooling as the aggregation function for $\mathcal{A}gg_h$ and $\mathcal{A}gg_s$, making $e_p$ reflect the average characteristics of the watched videos. 
More advanced techniques can adjust each video’s weight based on its relevance to the target video or the user’s interests.



\smallsection{Segment-wise Scoring}
The input representations are first processed through projection layers, each consisting of 3 sequential layers of LayerNorm and dropout. 
Similarly, the preference embeddings pass through the same structure of projection layers, ensuring alignment in a shared embedding space.
Next, a cross-attention layer uses the input representations as queries and the preference embeddings as keys and values to condition segment representations based on preferences.
Similar to \citep{lei2021detecting, narasimhan2021clip}, the attended output is then fed into a transformer encoder, which includes a multi-head self-attention layer and a feed forward network (FFN), to compute segment-wise saliency scores that capture the relevance of each segment based on the modeled preferences.

\smallsection{Saliency Loss}
We employ a contrastive saliency loss to ensure relevant clips receive higher saliency scores and irrelevant ones lower scores, enforcing a ranking based on user preferences.
Given a target video with segments $v^+$ (relevant) and $v^-$ (irrelevant), and their corresponding saliency scores $y^+$ and $y^-$, the loss is defined as:
\begin{equation}
\small
\mathcal{L}_{\text{saliency}} =\sum_{(v^+, v^-)} \max(0, \gamma - (y^+ - y^-))
\end{equation}
If the difference between $y^+$ and $y^-$ is smaller than $\gamma$, the loss function penalizes the model, encouraging it to assign a higher score to the relevant segment.

\section{Experiments}

\subsection{Experimental Settings}
In this section, we compare our task, \task (PV), with existing methods in video summarization (VS), highlight detection (HD), and moment retrieval (MR).
VS and HD use only the target video to select keyframes, while MR takes a natural language query as well to retrieve a matching temporal segment.
In contrast, PV uses both a user’s watch history and the target video to predict segment saliency scores.

\smallsection{Experimental Setup}
We split \dataset into training (70\%) and test (30\%) sets, ensuring a balanced ratio of videos across categories for content diversity.
For MR and query-focused VS, we generate text-based queries by extracting key phrases that capture the essence of the user’s watch history.
Additionally, we train on QVHighlights~\citep{lei2021detecting}, a widely-used dataset for HD and MR, and evaluate on \dataset.
This setup assesses the generalization ability of models across different datasets, though \datasets unique requirement for video history sequences limits direct cross-dataset generalization.

\smallsection{Baselines}
We evaluate recent state-of-the-art methods for personalized video highlighting.
For HD and MR, we include transformer-based models—SL-Module~\citep{xu2021cross}, UMT~\citep{liu2022umt}, and UVCOM~\citep{xiao2024bridging}—as well as DETR-based approaches: Moment-DETR~\citep{lei2021detecting}, QD-DETR~\citep{moon2023query}, and TR-DETR~\citep{sun2024tr}.
Note that SL-Module is applied only to HD.
For VS, we adapt CLIP-It~\citep{narasimhan2021clip} for generic and query-focused summaries, and VSL~\citep{Chen2024personalized} for personalized summaries. 
More details are provided in Appendix~\ref{apdx:baselines}.

\smallsection{Evaluation Metrics}
We evaluate model performance using standard metrics used in baselines~\citep{lei2021detecting,sul2023mr,moon2023query}.
For HD, we use mean average precision (mAP) and Hit@1 to assess ranking quality, with saliency score thresholds of 7 and 9 (out of 10).\footnote{\citet{liu2015multi, lei2021detecting, moon2023query} set the threshold as 4 out of 5.} 
For MR, we compute Recall@1 with IoU thresholds of 0.5 and 0.7 to measure temporal alignment accuracy. 
For VS, we use F1 score to assess the balance between precision and recall in segment selection.
Furthermore, since our task includes score prediction, we use Root Mean Square Error (RMSE) to evaluate segment relevance prediction accuracy.
More details on evaluation metrics can be found in Appendix~\ref{apdx:eval_metrics}.

\begin{table*}[tbp]
    \centering
    \resizebox{\linewidth}{!}{
    \begin{tabular}{l|cccccc|c}
        \toprule
        Method & RMSE $\downarrow$ & mAP & Hit1@7 & Hit1@9 & Recall1@0.5 & Recall1@0.7 & Improv. \\
        \midrule
        \rowcolor{gray!20}
        \multicolumn{8}{c}{QVHighlights} \\
        Moment-DETR~\citep{lei2021detecting} & 0.347 & 0.681 & 0.434 & 0.042 & \underline{0.370} & 0.205 & 20.7\% \\
        UMT~\citep{liu2022umt} & 0.527 & 0.547 & 0.409 & \underline{0.138} & 0.255 & 0.179 & 20.2\% \\
        QD-DETR~\citep{moon2023query} & 0.375 & 0.675 & 0.406 & 0.116 & 0.353 & 0.201 & 43.4\% \\
        UVCOM~\citep{xiao2024bridging} & 0.330 & 0.710 & 0.489 & 0.149 & 0.413 & 0.183 & 11.4\% \\
        TR-DETR~\citep{sun2024tr} & 0.400 & 0.660 & 0.352 & 0.105 & 0.359 & 0.195 & 58.1\% \\
        \midrule
        \rowcolor{gray!20}
        \multicolumn{8}{c}{\dataset} \\
        SL-Module~\citep{xu2021cross} & 0.517 & 0.568 & 0.385 & 0.085 & -- & -- & 96.1\% \\
        Moment-DETR~\citep{lei2021detecting} & \underline{0.339} & 0.705 & 0.432 & \underline{0.138} & 0.398 & 0.193 & 38.2\% \\
        UMT~\citep{liu2022umt} & 0.502 & \underline{0.732} & 0.429 & 0.132 & 0.320 & \underline{0.210} & 6.4\% \\
        QD-DETR~\citep{moon2023query} & 0.368 & 0.681 & \underline{0.456} & 0.120 & 0.365 & 0.196 & 38.2\% \\
        UVCOM~\citep{xiao2024bridging} & 0.350 & 0.700 & 0.441 & 0.146 & 0.357 & 0.154 & 13.7\% \\
        TR-DETR~\citep{sun2024tr} & 0.390 & 0.660 & 0.435 & 0.149 & 0.243 & 0.127 & 11.4\% \\
        \midrule
        \proposed & \textbf{0.301} & \textbf{0.766} & \textbf{0.507} & \textbf{0.166} & \textbf{0.452} & \textbf{0.245} & \\[0.001em]
        \bottomrule
    \end{tabular}
    }
    \caption{Performance comparison for HD and MR. Hit1@$k$ and Recall@@$\alpha$ are computed using saliency threshold $k$ and IoU threshold $\alpha$, respectively. Gray rows indicate training datasets. The best results are shown in \textbf{bold}, and the second-best are \underline{underlined}.}
    \label{tbl:performance}
\end{table*}

\subsection{Main Results}
Table~\ref{tbl:performance} shows that \proposed outperforms all baselines over evaluation metrics, highlighting its effectiveness in capturing user-specific preferences. 
This performance gain is primarily attributed to the incorporation of personalized understanding through watch histories.
As a generic approach, SL-Module effectively identifies informative segments but fails to reflect a user’s unique preferences. This limitation emphasizes the challenges of applying non-personalized methods in a personalized setting. 
While UMT, Moment-DETR, and QD-DETR, which leverage natural language queries, achieve better performance than generic baselines, they struggle to capture finer-grained moments. 
This might be because natural language queries provide a simplified representation of user intent compared to the richer contextual signals available in watch history.
On the other hand, UMT tends to show competitive performance with ours, since UMT and ours use additional audio sources;
this strongly indicates the importance of incorporation of multi-modal source.


\subsection{Additional Results on \dataseth}
\label{subsec:exp_dataseth}
\begin{wraptable}{r}{0.47\linewidth}
\vspace{-1.2em}
\centering
\resizebox{\linewidth}{!}{%
  \begin{tabular}{l|cccc}
    \toprule
    Method & RMSE & H1@7 & H1@9 & F1@0.5 \\
    \midrule
    Moment-DETR & \textbf{0.419} & 0.472 & 0.389 & 0.417 \\
    QD-DETR     & 0.446 & 0.444 & 0.361 & 0.385 \\
    TR-DETR     & 0.443 & 0.306 & 0.250 & 0.429 \\
    \midrule
    HiPher      & 0.427 & \textbf{0.486} & \textbf{0.400} & \textbf{0.624} \\
    \bottomrule
  \end{tabular}
}
\vspace{-0.5em}
\caption{Performance on \dataseth.}
\label{tab:hippo_h_eval}
\vspace{-1em}
\end{wraptable}
We further evaluate \proposed and the MR/HD baselines on \dataseth, a dataset collected from real users, to assess the practicality and generalizability of each method. 
As summarized in Table~\ref{tab:hippo_h_eval}, \proposed consistently outperforms the baselines across most metrics, showing strong robustness beyond simulated settings.
These findings highlight the effectiveness of \proposed in modeling nuanced viewing behaviors and suggest promising directions for future research in user-adaptive video understanding.

\subsection{Ablation Studies}
\begin{figure*}[tbp]
\centering
\begin{minipage}[t]{0.67\textwidth}  
    \vspace{0pt} 
    \centering
    \small
    \begin{tabular}{l|c|cc}
        \toprule
        Method & Query Type & F1@5 & F1@7 \\
        \midrule
        Clip-It~\citep{narasimhan2021clip} & - & 0.564 & 0.211 \\
        Clip-It~\citep{narasimhan2021clip} & phrase & 0.566 & 0.230 \\
        Clip-It~\citep{narasimhan2021clip} & sentence & \underline{0.658} & \underline{0.234} \\
        VSL~\citep{Chen2024personalized} & genre & 0.466 & 0.187 \\
        \midrule
        \textbf{\proposed} & history & \textbf{0.726} & \textbf{0.486} \\
        \bottomrule
    \end{tabular}
    \captionof{table}{Performance comparison for video summarization. CLIP-It supports both generic (query-free) and query-focused summarization, while VSL provides personalized summarization based on preference queries (genres).}
    \label{tbl:ablation}
\end{minipage}%
\hfill
\begin{minipage}[t]{0.31\textwidth} 
    \vspace{-0.5em} 
    \centering
    \includegraphics[width=\linewidth]{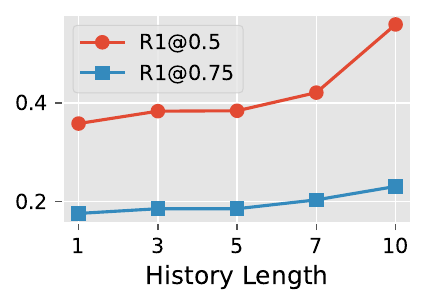}
    \vspace{-1.78em}
    \captionof{figure}{Performance comparison with varying numbers of history videos used for preference modeling.}
    \label{fig:history_length}
\end{minipage}
\end{figure*}

\smallsection{Query Type}
Table~\ref{tbl:ablation} reports summarization accuracy (F1 score at different thresholds) across various preference contexts: simple word-level queries, sentence-level descriptions, and user watch histories. 
\proposed performs the best when leveraging history, significantly outperforming word- and sentence-based representations.
These results emphasize the critical role of history-driven preference modeling for effective personalized video highlighting.


\smallsection{History Length}
We hypothesize that user preferences, which are often highly specific, can be more accurately captured by analyzing longer watch histories, as they reveal consistent patterns across a sequence.
To validate this, we conduct an ablation study by varying the number of watched videos (i.e., the length of the watch history) provided to the preference modeling module. 
As shown in Figure~\ref{fig:history_length}, performance improves as more history videos are included.
These results suggest that longer histories help surface repetitive cues, leading to more effective preference modeling and improved personalization in video engagement.

\begin{wraptable}{r}{0.37\linewidth} 
    \centering
    \resizebox{\linewidth}{!}{%
    \begin{tabular}{l|ccc}
        \toprule
        Method & mAP & H1@7 & R1@0.5 \\
        \midrule
        \proposed-V & 0.67 & 0.12 & 0.32 \\
        \proposed-T & 0.74 & 0.15 & 0.39 \\
        \midrule
        \proposed & \textbf{0.77} & \textbf{0.17} & \textbf{0.45} \\
        \bottomrule
    \end{tabular}
    }
    \caption{Ablation results on different input modalities.}
    \label{tbl:modality-ablation}
    \vspace{-1em}
\end{wraptable}
\textbf{Input Modalities. }
We conduct an ablation study to evaluate the contributions of visual and textual features. As shown in Table~\ref{tbl:modality-ablation}, using a single modality leads to reduced performance, with textual features (\proposed-V) being more informative than visual ones (\proposed-T). Combining both modalities (\proposed) achieves the best results, demonstrating the effectiveness of our multimodal approach in capturing fine-grained user preferences for \task. This is particularly important given the diversity of \dataset, which reflects the range of real-world videos, including both visually rich and narrative-only content.

\begin{figure*}[t]
    \centering    
    \includegraphics[width=\linewidth]{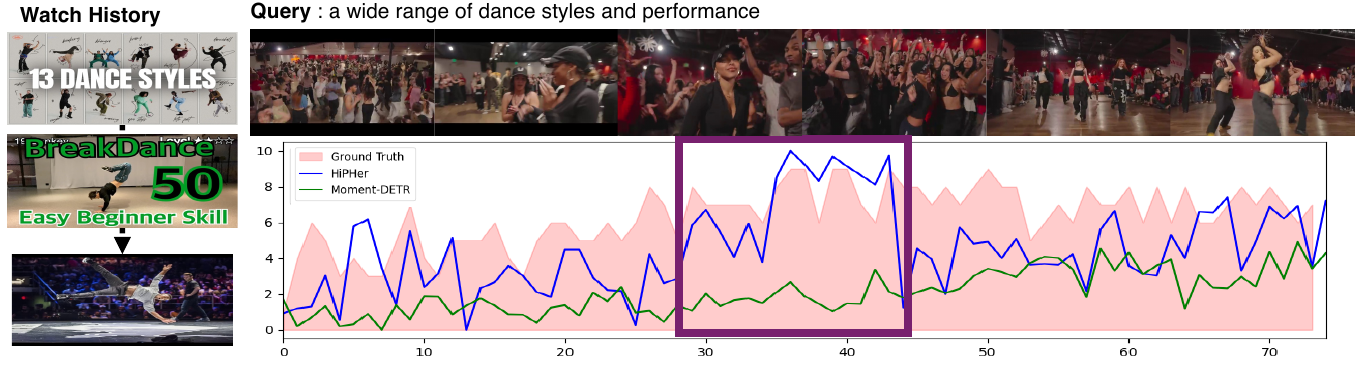}
    \caption{Case study on saliency (preference) scores between \proposed and Moment-DETR.}
    \label{fig:case-study}
\end{figure*}

\smallsection{Case Study}
We present a case study that qualitatively compares the saliency (preference) scores of \proposed and Moment-DETR.
Figure~\ref{fig:case-study} visualizes the segment-wise scores within a target video, contrasting a history-focused embedding approach with a query-based method.
Overall, the blue line (\proposed) closely aligns with the ground truth scores, while the green line (Moment-DETR) sometimes shows notable discrepancies (highlighted in the purple box), indicating that the history-driven embedding provides richer contextual information for personalization to a text query.
Additional case studies are provided in Appendix~\ref{apdx:case_studies}.

\section{Conclusion}
Motivated by the need to tailor video content to individual preferences in real-world scenarios, we introduce \task, a novel task that leverages user watch history to highlight relevant video segments. 
We also present \dataset, a unique dataset generated through LLM-based user simulation, which includes user watch histories and personalized saliency scores.
Through comprehensive experiments, we demonstrate that history-driven preference modeling significantly improves performance, surpassing existing methods based on generic or text-based queries. 
Our findings emphasize the value of integrating user-specific preferences and history for more effective video content delivery, offering promising directions for future advancements in personalized video experiences.

\section*{Acknowledgement}
This work was supported by the IITP grants funded by the Korea government (MSIT) (No. RS-2020- II201361; RS-2024-00457882, AI Research Hub Project), and the NRF grant funded by the Korea government (MSIT) (No. RS-2025-00560295).

\bibliography{colm2025_conference}

\begin{thebibliography}{56}
\providecommand{\natexlab}[1]{#1}
\providecommand{\url}[1]{\texttt{#1}}
\expandafter\ifx\csname urlstyle\endcsname\relax
  \providecommand{\doi}[1]{doi: #1}\else
  \providecommand{\doi}{doi: \begingroup \urlstyle{rm}\Url}\fi

\bibitem[Achiam et~al.(2023)Achiam, Adler, Agarwal, Ahmad, Akkaya, Aleman, Almeida, Altenschmidt, Altman, Anadkat, et~al.]{achiam2023gpt}
Josh Achiam, Steven Adler, Sandhini Agarwal, Lama Ahmad, Ilge Akkaya, Florencia~Leoni Aleman, Diogo Almeida, Janko Altenschmidt, Sam Altman, Shyamal Anadkat, et~al.
\newblock Gpt-4 technical report.
\newblock \emph{arXiv preprint arXiv:2303.08774}, 2023.

\bibitem[Apostolidis et~al.(2021)Apostolidis, Adamantidou, Metsai, Mezaris, and Patras]{apostolidis2021video}
Evlampios Apostolidis, Eleni Adamantidou, Alexandros~I Metsai, Vasileios Mezaris, and Ioannis Patras.
\newblock Video summarization using deep neural networks: A survey.
\newblock \emph{Proceedings of the IEEE}, 109\penalty0 (11):\penalty0 1838--1863, 2021.

\bibitem[Argaw et~al.(2024{\natexlab{a}})Argaw, Soldan, Pardo, Zhao, Heilbron, Chung, and Ghanem]{argaw2024towards}
Dawit~Mureja Argaw, Mattia Soldan, Alejandro Pardo, Chen Zhao, Fabian~Caba Heilbron, Joon~Son Chung, and Bernard Ghanem.
\newblock Towards automated movie trailer generation.
\newblock In \emph{Proceedings of the IEEE/CVF Conference on Computer Vision and Pattern Recognition}, pp.\  7445--7454, 2024{\natexlab{a}}.

\bibitem[Argaw et~al.(2024{\natexlab{b}})Argaw, Yoon, Heilbron, Deilamsalehy, Bui, Wang, Dernoncourt, and Chung]{argaw2024scaling}
Dawit~Mureja Argaw, Seunghyun Yoon, Fabian~Caba Heilbron, Hanieh Deilamsalehy, Trung Bui, Zhaowen Wang, Franck Dernoncourt, and Joon~Son Chung.
\newblock Scaling up video summarization pretraining with large language models.
\newblock In \emph{Proceedings of the IEEE/CVF Conference on Computer Vision and Pattern Recognition}, pp.\  8332--8341, 2024{\natexlab{b}}.

\bibitem[Bao et~al.(2023)Bao, Zhao, Teng, Yang, and Zhang]{bao2023fast}
Guangsheng Bao, Yanbin Zhao, Zhiyang Teng, Linyi Yang, and Yue Zhang.
\newblock Fast-detectgpt: Efficient zero-shot detection of machine-generated text via conditional probability curvature.
\newblock \emph{arXiv preprint arXiv:2310.05130}, 2023.

\bibitem[Carion et~al.(2020)Carion, Massa, Synnaeve, Usunier, Kirillov, and Zagoruyko]{carion2020end}
Nicolas Carion, Francisco Massa, Gabriel Synnaeve, Nicolas Usunier, Alexander Kirillov, and Sergey Zagoruyko.
\newblock End-to-end object detection with transformers.
\newblock In \emph{European conference on computer vision}, pp.\  213--229. Springer, 2020.

\bibitem[Chen et~al.(2024)Chen, Zhao, and Zhu]{Chen2024personalized}
Brian Chen, Xiangyuan Zhao, and Yingnan Zhu.
\newblock Personalized video summarization by multimodal video understanding.
\newblock In \emph{CIKM}, 2024.

\bibitem[Chiang et~al.(2024)Chiang, Zheng, Sheng, Angelopoulos, Li, Li, Zhu, Zhang, Jordan, Gonzalez, et~al.]{chiang2024chatbot}
Wei-Lin Chiang, Lianmin Zheng, Ying Sheng, Anastasios~Nikolas Angelopoulos, Tianle Li, Dacheng Li, Banghua Zhu, Hao Zhang, Michael Jordan, Joseph~E Gonzalez, et~al.
\newblock Chatbot arena: An open platform for evaluating llms by human preference.
\newblock In \emph{Forty-first International Conference on Machine Learning}, 2024.

\bibitem[Gao et~al.(2017)Gao, Sun, Yang, and Nevatia]{gao2017tall}
Jiyang Gao, Chen Sun, Zhenheng Yang, and Ram Nevatia.
\newblock Tall: Temporal activity localization via language query.
\newblock In \emph{Proceedings of the IEEE international conference on computer vision}, pp.\  5267--5275, 2017.

\bibitem[Gygli et~al.(2014)Gygli, Grabner, Riemenschneider, and Van~Gool]{gygli2014creating}
Michael Gygli, Helmut Grabner, Hayko Riemenschneider, and Luc Van~Gool.
\newblock Creating summaries from user videos.
\newblock In \emph{Computer Vision--ECCV 2014: 13th European Conference, Zurich, Switzerland, September 6-12, 2014, Proceedings, Part VII 13}, pp.\  505--520. Springer, 2014.

\bibitem[Gygli et~al.(2016)Gygli, Song, and Cao]{gygli2016video2gif}
Michael Gygli, Yale Song, and Liangliang Cao.
\newblock Video2gif: Automatic generation of animated gifs from video.
\newblock In \emph{Proceedings of the IEEE conference on computer vision and pattern recognition}, pp.\  1001--1009, 2016.

\bibitem[Han et~al.(2023)Han, Yang, Chang, and Wang]{han2023shot2story20k}
Mingfei Han, Linjie Yang, Xiaojun Chang, and Heng Wang.
\newblock Shot2story20k: A new benchmark for comprehensive understanding of multi-shot videos.
\newblock \emph{arXiv preprint arXiv:2312.10300}, 2023.

\bibitem[Hua et~al.(2024)Hua, Tang, Xu, and Luo]{hua2024v2xum}
Hang Hua, Yunlong Tang, Chenliang Xu, and Jiebo Luo.
\newblock V2xum-llm: Cross-modal video summarization with temporal prompt instruction tuning.
\newblock \emph{arXiv preprint arXiv:2404.12353}, 2024.

\bibitem[Huang et~al.(2020)Huang, Xiong, Rao, Wang, and Lin]{huang2020movienet}
Qingqiu Huang, Yu~Xiong, Anyi Rao, Jiaze Wang, and Dahua Lin.
\newblock Movienet: A holistic dataset for movie understanding.
\newblock In \emph{Computer Vision--ECCV 2020: 16th European Conference, Glasgow, UK, August 23--28, 2020, Proceedings, Part IV 16}, pp.\  709--727. Springer, 2020.

\bibitem[Ji et~al.(2019)Ji, Xiong, Pang, and Li]{ji2019video}
Zhong Ji, Kailin Xiong, Yanwei Pang, and Xuelong Li.
\newblock Video summarization with attention-based encoder--decoder networks.
\newblock \emph{IEEE Transactions on Circuits and Systems for Video Technology}, 30\penalty0 (6):\penalty0 1709--1717, 2019.

\bibitem[Kamath et~al.(2021)Kamath, Singh, LeCun, Synnaeve, Misra, and Carion]{kamath2021mdetr}
Aishwarya Kamath, Mannat Singh, Yann LeCun, Gabriel Synnaeve, Ishan Misra, and Nicolas Carion.
\newblock Mdetr-modulated detection for end-to-end multi-modal understanding.
\newblock In \emph{Proceedings of the IEEE/CVF international conference on computer vision}, pp.\  1780--1790, 2021.

\bibitem[Kang \& McAuley(2018)Kang and McAuley]{kang2018self}
Wang-Cheng Kang and Julian McAuley.
\newblock Self-attentive sequential recommendation.
\newblock In \emph{2018 IEEE international conference on data mining (ICDM)}, pp.\  197--206. IEEE, 2018.

\bibitem[Lei et~al.(2020)Lei, Yu, Berg, and Bansal]{lei2020tvr}
Jie Lei, Licheng Yu, Tamara~L Berg, and Mohit Bansal.
\newblock Tvr: A large-scale dataset for video-subtitle moment retrieval.
\newblock In \emph{Computer Vision--ECCV 2020: 16th European Conference, Glasgow, UK, August 23--28, 2020, Proceedings, Part XXI 16}, pp.\  447--463. Springer, 2020.

\bibitem[Lei et~al.(2021)Lei, Berg, and Bansal]{lei2021detecting}
Jie Lei, Tamara~L Berg, and Mohit Bansal.
\newblock Detecting moments and highlights in videos via natural language queries.
\newblock \emph{Advances in Neural Information Processing Systems}, 34:\penalty0 11846--11858, 2021.

\bibitem[Li et~al.(2020)Li, Chen, Cheng, Gan, Yu, and Liu]{li2020hero}
Linjie Li, Yen-Chun Chen, Yu~Cheng, Zhe Gan, Licheng Yu, and Jingjing Liu.
\newblock Hero: Hierarchical encoder for video+ language omni-representation pre-training.
\newblock \emph{arXiv preprint arXiv:2005.00200}, 2020.

\bibitem[Lin et~al.(2023)Lin, Zhang, Chen, Pramanick, Gao, Wang, Yan, and Shou]{lin2023univtg}
Kevin~Qinghong Lin, Pengchuan Zhang, Joya Chen, Shraman Pramanick, Difei Gao, Alex~Jinpeng Wang, Rui Yan, and Mike~Zheng Shou.
\newblock Univtg: Towards unified video-language temporal grounding.
\newblock In \emph{Proceedings of the IEEE/CVF International Conference on Computer Vision}, pp.\  2794--2804, 2023.

\bibitem[Liu et~al.(2024)Liu, Li, Wu, and Lee]{liu2024visual}
Haotian Liu, Chunyuan Li, Qingyang Wu, and Yong~Jae Lee.
\newblock Visual instruction tuning.
\newblock \emph{Advances in neural information processing systems}, 36, 2024.

\bibitem[Liu et~al.(2018)Liu, Wang, Nie, He, Chen, and Chua]{liu2018attentive}
Meng Liu, Xiang Wang, Liqiang Nie, Xiangnan He, Baoquan Chen, and Tat-Seng Chua.
\newblock Attentive moment retrieval in videos.
\newblock In \emph{The 41st international ACM SIGIR conference on research \& development in information retrieval}, pp.\  15--24, 2018.

\bibitem[Liu et~al.(2015)Liu, Mei, Zhang, Che, and Luo]{liu2015multi}
Wu~Liu, Tao Mei, Yongdong Zhang, Cherry Che, and Jiebo Luo.
\newblock Multi-task deep visual-semantic embedding for video thumbnail selection.
\newblock In \emph{Proceedings of the IEEE conference on computer vision and pattern recognition}, pp.\  3707--3715, 2015.

\bibitem[Liu et~al.(2022)Liu, Li, Wu, Chen, Shan, and Qie]{liu2022umt}
Ye~Liu, Siyuan Li, Yang Wu, Chang-Wen Chen, Ying Shan, and Xiaohu Qie.
\newblock Umt: Unified multi-modal transformers for joint video moment retrieval and highlight detection.
\newblock In \emph{Proceedings of the IEEE/CVF Conference on Computer Vision and Pattern Recognition}, pp.\  3042--3051, 2022.

\bibitem[Lu et~al.(2019)Lu, Chen, Tan, Li, and Xiao]{lu2019debug}
Chujie Lu, Long Chen, Chilie Tan, Xiaolin Li, and Jun Xiao.
\newblock Debug: A dense bottom-up grounding approach for natural language video localization.
\newblock In \emph{Proceedings of the 2019 Conference on Empirical Methods in Natural Language Processing and the 9th International Joint Conference on Natural Language Processing (EMNLP-IJCNLP)}, pp.\  5144--5153, 2019.

\bibitem[Luo et~al.(2025)Luo, Wu, Li, Ma, Kankanhalli, and Li]{luo2025videoautoarena}
Ziyang Luo, Haoning Wu, Dongxu Li, Jing Ma, Mohan Kankanhalli, and Junnan Li.
\newblock Videoautoarena: An automated arena for evaluating large multimodal models in video analysis through user simulation.
\newblock In \emph{Proceedings of the Computer Vision and Pattern Recognition Conference}, pp.\  8461--8474, 2025.

\bibitem[Miech et~al.(2019)Miech, Zhukov, Alayrac, Tapaswi, Laptev, and Sivic]{miech2019howto100m}
Antoine Miech, Dimitri Zhukov, Jean-Baptiste Alayrac, Makarand Tapaswi, Ivan Laptev, and Josef Sivic.
\newblock Howto100m: Learning a text-video embedding by watching hundred million narrated video clips.
\newblock In \emph{Proceedings of the IEEE/CVF international conference on computer vision}, pp.\  2630--2640, 2019.

\bibitem[Mitchell et~al.(2023)Mitchell, Lee, Khazatsky, Manning, and Finn]{mitchell2023detectgpt}
Eric Mitchell, Yoonho Lee, Alexander Khazatsky, Christopher~D Manning, and Chelsea Finn.
\newblock Detectgpt: Zero-shot machine-generated text detection using probability curvature.
\newblock In \emph{International Conference on Machine Learning}, pp.\  24950--24962. PMLR, 2023.

\bibitem[Moon et~al.(2023)Moon, Hyun, Park, Park, and Heo]{moon2023query}
WonJun Moon, Sangeek Hyun, SangUk Park, Dongchan Park, and Jae-Pil Heo.
\newblock Query-dependent video representation for moment retrieval and highlight detection.
\newblock In \emph{Proceedings of the IEEE/CVF Conference on Computer Vision and Pattern Recognition}, pp.\  23023--23033, 2023.

\bibitem[Narasimhan et~al.(2021)Narasimhan, Rohrbach, and Darrell]{narasimhan2021clip}
Medhini Narasimhan, Anna Rohrbach, and Trevor Darrell.
\newblock Clip-it! language-guided video summarization.
\newblock \emph{Advances in neural information processing systems}, 34:\penalty0 13988--14000, 2021.

\bibitem[Park et~al.(2020)Park, Lee, Kim, and Sohn]{park2020sumgraph}
Jungin Park, Jiyoung Lee, Ig-Jae Kim, and Kwanghoon Sohn.
\newblock Sumgraph: Video summarization via recursive graph modeling.
\newblock In \emph{Computer Vision--ECCV 2020: 16th European Conference, Glasgow, UK, August 23--28, 2020, Proceedings, Part XXV 16}, pp.\  647--663. Springer, 2020.

\bibitem[Qiu et~al.(2024)Qiu, Zhu, Han, Kumar, Mittal, Jin, Yang, Li, Wang, Zhao, et~al.]{qiu2024mmsum}
Jielin Qiu, Jiacheng Zhu, William Han, Aditesh Kumar, Karthik Mittal, Claire Jin, Zhengyuan Yang, Linjie Li, Jianfeng Wang, Ding Zhao, et~al.
\newblock Mmsum: A dataset for multimodal summarization and thumbnail generation of videos.
\newblock In \emph{Proceedings of the IEEE/CVF Conference on Computer Vision and Pattern Recognition}, pp.\  21909--21921, 2024.

\bibitem[Radford et~al.(2021)Radford, Kim, Hallacy, Ramesh, Goh, Agarwal, Sastry, Askell, Mishkin, Clark, et~al.]{radford2021learning}
Alec Radford, Jong~Wook Kim, Chris Hallacy, Aditya Ramesh, Gabriel Goh, Sandhini Agarwal, Girish Sastry, Amanda Askell, Pamela Mishkin, Jack Clark, et~al.
\newblock Learning transferable visual models from natural language supervision.
\newblock In \emph{International conference on machine learning}, pp.\  8748--8763. PMLR, 2021.

\bibitem[Rendle et~al.(2009)Rendle, Freudenthaler, Gantner, and Schmidt-Thieme]{rendle2009bpr}
Steffen Rendle, Christoph Freudenthaler, Zeno Gantner, and Lars Schmidt-Thieme.
\newblock Bpr: Bayesian personalized ranking from implicit feedback.
\newblock In \emph{Proceedings of the Twenty-Fifth Conference on Uncertainty in Artificial Intelligence}, pp.\  452--461, 2009.

\bibitem[Rochan et~al.(2020)Rochan, Krishna~Reddy, Ye, and Wang]{rochan2020adaptive}
Mrigank Rochan, Mahesh~Kumar Krishna~Reddy, Linwei Ye, and Yang Wang.
\newblock Adaptive video highlight detection by learning from user history.
\newblock In \emph{European conference on computer vision}, pp.\  261--278. Springer, 2020.

\bibitem[Sharghi et~al.(2016)Sharghi, Gong, and Shah]{sharghi2016query}
Aidean Sharghi, Boqing Gong, and Mubarak Shah.
\newblock Query-focused extractive video summarization.
\newblock In \emph{Computer Vision--ECCV 2016: 14th European Conference, Amsterdam, The Netherlands, October 11-14, 2016, Proceedings, Part VIII 14}, pp.\  3--19. Springer, 2016.

\bibitem[Sharghi et~al.(2017)Sharghi, Laurel, and Gong]{sharghi2017query}
Aidean Sharghi, Jacob~S Laurel, and Boqing Gong.
\newblock Query-focused video summarization: Dataset, evaluation, and a memory network based approach.
\newblock In \emph{Proceedings of the IEEE conference on computer vision and pattern recognition}, pp.\  4788--4797, 2017.

\bibitem[Song et~al.(2015)Song, Vallmitjana, Stent, and Jaimes]{song2015tvsum}
Yale Song, Jordi Vallmitjana, Amanda Stent, and Alejandro Jaimes.
\newblock Tvsum: Summarizing web videos using titles.
\newblock In \emph{Proceedings of the IEEE conference on computer vision and pattern recognition}, pp.\  5179--5187, 2015.

\bibitem[Song et~al.(2016)Song, Redi, Vallmitjana, and Jaimes]{song2016click}
Yale Song, Miriam Redi, Jordi Vallmitjana, and Alejandro Jaimes.
\newblock To click or not to click: Automatic selection of beautiful thumbnails from videos.
\newblock In \emph{Proceedings of the 25th ACM international on conference on information and knowledge management}, pp.\  659--668, 2016.

\bibitem[Sul et~al.(2023)Sul, Han, and Lee]{sul2023mr}
Jinhwan Sul, Jihoon Han, and Joonseok Lee.
\newblock Mr. hisum: A large-scale dataset for video highlight detection and summarization.
\newblock \emph{Advances in Neural Information Processing Systems}, 36:\penalty0 40542--40555, 2023.

\bibitem[Sun et~al.(2024)Sun, Zhou, Chen, and Xie]{sun2024tr}
Hao Sun, Mingyao Zhou, Wenjing Chen, and Wei Xie.
\newblock Tr-detr: Task-reciprocal transformer for joint moment retrieval and highlight detection.
\newblock In \emph{Proceedings of the AAAI Conference on Artificial Intelligence}, volume~38, pp.\  4998--5007, 2024.

\bibitem[Sun et~al.(2014)Sun, Farhadi, and Seitz]{sun2014ranking}
Min Sun, Ali Farhadi, and Steve Seitz.
\newblock Ranking domain-specific highlights by analyzing edited videos.
\newblock In \emph{Computer Vision--ECCV 2014: 13th European Conference, Zurich, Switzerland, September 6-12, 2014, Proceedings, Part I 13}, pp.\  787--802. Springer, 2014.

\bibitem[Vasudevan et~al.(2017)Vasudevan, Gygli, Volokitin, and Van~Gool]{vasudevan2017query}
Arun~Balajee Vasudevan, Michael Gygli, Anna Volokitin, and Luc Van~Gool.
\newblock Query-adaptive video summarization via quality-aware relevance estimation.
\newblock In \emph{Proceedings of the 25th ACM international conference on Multimedia}, pp.\  582--590, 2017.

\bibitem[Wang et~al.(2024)Wang, Zhang, Zohar, and Yeung-Levy]{wang2024videoagent}
Xiaohan Wang, Yuhui Zhang, Orr Zohar, and Serena Yeung-Levy.
\newblock Videoagent: Long-form video understanding with large language model as agent.
\newblock In \emph{European Conference on Computer Vision}, pp.\  58--76. Springer, 2024.

\bibitem[Xiao et~al.(2020{\natexlab{a}})Xiao, Zhao, Zhang, Guan, and Cai]{xiao2020query}
Shuwen Xiao, Zhou Zhao, Zijian Zhang, Ziyu Guan, and Deng Cai.
\newblock Query-biased self-attentive network for query-focused video summarization.
\newblock \emph{IEEE Transactions on Image Processing}, 29:\penalty0 5889--5899, 2020{\natexlab{a}}.

\bibitem[Xiao et~al.(2020{\natexlab{b}})Xiao, Zhao, Zhang, Yan, and Yang]{xiao2020convolutional}
Shuwen Xiao, Zhou Zhao, Zijian Zhang, Xiaohui Yan, and Min Yang.
\newblock Convolutional hierarchical attention network for query-focused video summarization.
\newblock In \emph{Proceedings of the AAAI conference on artificial intelligence}, volume~34, pp.\  12426--12433, 2020{\natexlab{b}}.

\bibitem[Xiao et~al.(2024)Xiao, Luo, Liu, Ma, Bian, Ji, Yang, and Li]{xiao2024bridging}
Yicheng Xiao, Zhuoyan Luo, Yong Liu, Yue Ma, Hengwei Bian, Yatai Ji, Yujiu Yang, and Xiu Li.
\newblock Bridging the gap: A unified video comprehension framework for moment retrieval and highlight detection.
\newblock In \emph{Proceedings of the IEEE/CVF Conference on Computer Vision and Pattern Recognition}, pp.\  18709--18719, 2024.

\bibitem[Xu et~al.(2021)Xu, Wang, Ni, Zhu, Sun, and Wang]{xu2021cross}
Minghao Xu, Hang Wang, Bingbing Ni, Riheng Zhu, Zhenbang Sun, and Changhu Wang.
\newblock Cross-category video highlight detection via set-based learning.
\newblock In \emph{Proceedings of the IEEE/CVF International Conference on Computer Vision}, pp.\  7970--7979, 2021.

\bibitem[Xu et~al.(2024)Xu, Sun, Zhai, Jia, and Du]{xu2024mh}
Yifang Xu, Yunzhuo Sun, Benxiang Zhai, Youyao Jia, and Sidan Du.
\newblock Mh-detr: Video moment and highlight detection with cross-modal transformer.
\newblock In \emph{2024 International Joint Conference on Neural Networks (IJCNN)}, pp.\  1--8. IEEE, 2024.

\bibitem[Yao et~al.(2016)Yao, Mei, and Rui]{yao2016highlight}
Ting Yao, Tao Mei, and Yong Rui.
\newblock Highlight detection with pairwise deep ranking for first-person video summarization.
\newblock In \emph{Proceedings of the IEEE conference on computer vision and pattern recognition}, pp.\  982--990, 2016.

\bibitem[Yuan et~al.(2019)Yuan, Ma, Wang, Liu, and Zhu]{yuan2019semantic}
Yitian Yuan, Lin Ma, Jingwen Wang, Wei Liu, and Wenwu Zhu.
\newblock Semantic conditioned dynamic modulation for temporal sentence grounding in videos.
\newblock \emph{Advances in Neural Information Processing Systems}, 32, 2019.

\bibitem[Zala et~al.(2023)Zala, Cho, Kottur, Chen, Oguz, Mehdad, and Bansal]{zala2023hierarchical}
Abhay Zala, Jaemin Cho, Satwik Kottur, Xilun Chen, Barlas Oguz, Yashar Mehdad, and Mohit Bansal.
\newblock Hierarchical video-moment retrieval and step-captioning.
\newblock In \emph{Proceedings of the IEEE/CVF Conference on Computer Vision and Pattern Recognition}, pp.\  23056--23065, 2023.

\bibitem[Zeng et~al.(2022)Zeng, Cao, Lu, Zhang, Xu, and Qin]{zeng2022moment}
Yawen Zeng, Da~Cao, Shaofei Lu, Hanling Zhang, Jiao Xu, and Zheng Qin.
\newblock Moment is important: Language-based video moment retrieval via adversarial learning.
\newblock \emph{ACM Transactions on Multimedia Computing, Communications, and Applications (TOMM)}, 18\penalty0 (2):\penalty0 1--21, 2022.

\bibitem[Zhang et~al.(2020)Zhang, Sun, Jing, and Zhou]{zhang2020span}
Hao Zhang, Aixin Sun, Wei Jing, and Joey~Tianyi Zhou.
\newblock Span-based localizing network for natural language video localization.
\newblock \emph{arXiv preprint arXiv:2004.13931}, 2020.

\bibitem[Zhou et~al.(2018)Zhou, Xu, and Corso]{zhou2018towards}
Luowei Zhou, Chenliang Xu, and Jason Corso.
\newblock Towards automatic learning of procedures from web instructional videos.
\newblock In \emph{Proceedings of the AAAI Conference on Artificial Intelligence}, volume~32, 2018.

\end{thebibliography}
\bibliographystyle{colm2025_conference}

\clearpage
\appendix
\section{Data Collection Process}

\subsection{User Initial Profile}
\label{apdx:user_init_profile}

For the initial preference seeds, we adopt 170 topic and sub-topic pairs adapted from the categorization proposed by \citet{qiu2024mmsum}, comprising 17 main categories each subdivided into 10 specific subcategories (see Table~\ref{tbl:categories}). Additionally, each pair is further annotated with a sentiment-based variable (\textit{intent}), represented by one of four features: \textbf{amusing}, \textbf{emotional}, \textbf{informative}, and \textbf{recent news}. 

\begin{table*}[thbp]
    \centering
    \small
    \renewcommand{\arraystretch}{1.1}
    \begin{tabular}{l|p{0.75\textwidth}}
        \toprule
        \textbf{Topics} & \textbf{Sub-topics} \\ 
        \midrule
        Animals &  Dog, Wildlife, Cat, Fish, Birds, Insect, Snakes, Pet, Amphibians, Reptile \\
        Education &  School, Club, Teacher, Speaking, Listening, Writing, Presentation, Math, Computer, Teamwork \\
        Health &  Mental, Injury, Medication, Digestive health, Dental, Optical, Reproductive, Skin, Brain health, Cardiac \\
        Travel &  Museum, Park, Sea, Beach, Mountain, Lake, Hotel, Resort, Camping, Hiking \\
        Movies &  Action movie, Comedy, Romance, Science fiction, Horror, Drama, Cartoon, Documentary, Adventure, Crime \\
        Cooking &  Broiling, Grilling, Roasting, Baking, Sauteing, Boiling, Steaming, Poaching, Simmering, Stewing \\
        Job &  Manager, Researcher, Chef, Police, Lawyer, Salesman, Mechanic, Banker, Doctor, Waiter \\
        Electronics &  Laptop, TV, Phone, Software, Internet, Camera, Audio, Headphone, Hardware, Monitor \\
        Art &  Crafts, Photography, Painting, Collection, Drawing, Digital art, Sculpting, Pottery, Glass craft, Calligraphy \\
        Personal Style &  Grooming, Fashion, Personal Hygiene, Tattoos, Scarf, Hair Style, Makeup, Dressing, Tie, Formal \\
        Clothes &  Sweater, Jeans, Shirt, Socks, Coat, Pants, Hat, Gloves, Dress, Shoes \\
        Sports &  Outdoor recreation, Team sports, Tennis, Football, Basketball, Climbing, Skiing, Swimming, Fishing, Yoga \\
        House &  Building, Garden, Pool, Bathroom, Bedroom, Kitchen, Repairment, Moving, Decoration, Furniture \\
        Food &  Fruit, Vegetable, Drink, Meat, Seafood, Snacks, Dessert, Breakfast, Lunch, Dinner \\
        Holiday &  Halloween, Christmas, Labor day, Thanksgiving, Valentine's day, Mother’s day, Birthday, National day, New year, Father’s day \\
        Transportation &  Car, Train, Bus, Boat, Bike, Airplane, Motorcycle, Truck, Trailer, Scooter \\
        Hobbies &  Dancing, Singing, Playing cards, Reading, Chess, Board games, Team games, Volunteer work, Instrument, Exercise \\
        \bottomrule
    \end{tabular}
    \caption{Topic and sub-topic pairs.}
    \label{tbl:categories}
\end{table*}

\subsection{YouTube Crawling}
\label{apdx:youtube_crawling}

\begin{figure}[h]
    \centering    
    \includegraphics[width=0.7\linewidth]{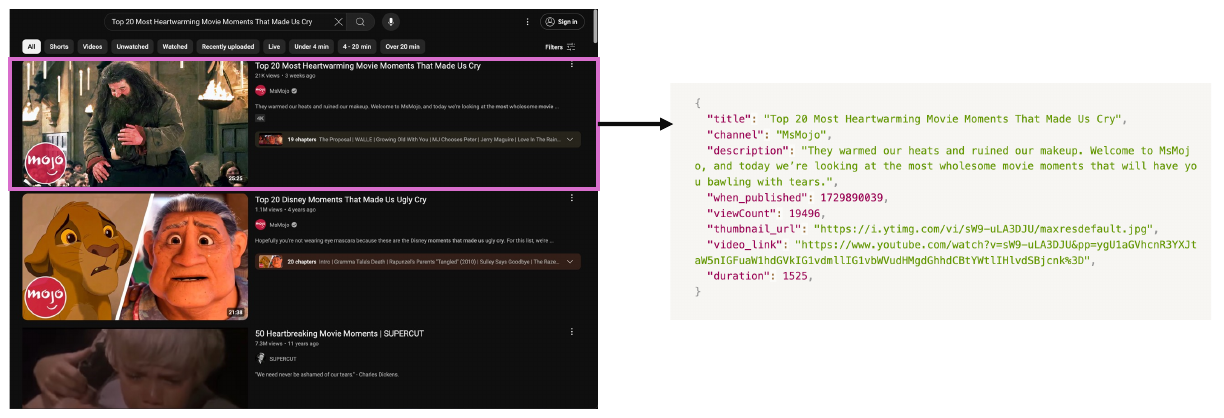} 
    \caption{The process of converting YouTube video metadata into a structured JSON format.}
    \label{suppl:video-json}
\end{figure}

We aim to replicate the environment where real users interact with YouTube to ensure realism in the LLM-based user simulator. To achieve this, as shown in Figure~\ref{suppl:video-json}, we provide the simulator with the exact metadata as displayed on the YouTube website, including the video title, channel name, description, view count, publication date, thumbnail URL, video link, and duration.

\subsection{Prompts}
\label{apdx:prompts}

\subsubsection{Video Candidate Retrieval}
Table~\ref{tbl:prompt-video-retrieve} shows the prompt used by the LLM-based user simulator to determine whether to explore related videos or generate a new search query.

\subsubsection{Video Engagement}
Table~\ref{tbl:prompt-video-selec} presents the prompt for selecting the most and least preferred videos from the candidate pool, while Table~\ref{tbl:prompt-video-watch} shows the prompt for engaging with the most preferred video.

\subsubsection{Preference Update}
Table~\ref{tbl:prompt-preference-update} shows the prompt used to update long-term preferences after interaction with the newly watched video.

\subsubsection{Saliency Scoring Annotation}
Table~\ref{tbl:prompt-score} shows the prompt used to assign saliency scores to each segment in the video based on the provided preference information.

\begin{table}[h]
\centering
\small
\resizebox{0.99\linewidth}{!}{%
\begin{tabular}{p{\linewidth}}
\toprule
\textbf{Prompt for Video Candidate Retrieval} \\
\midrule
You are finding \{intent\} videos about \{search query\}. \\[1em]

You have watched the following videos: \\
\{watch history\} \\[1em]

Your preferences have previously been defined as: \\
\{preference\} \\[1em]

For reference, current related videos are: \\
\{related videos\} \\[1em]

Now, decide whether to: \\
Explore the current query further by watching related videos. \\
Search for a new query to broaden your interest. \\[1em]

If you search for a new query, suggest one based on your interests, preferences, and history. \\[1em]

Answer Format: \\
Decision: ["Explore" or "Search for a new query"] \\
New query: [new query suggestion if "Search for a new query"] \\
\bottomrule
\end{tabular}%
}
\caption{Instructions for an LLM-based user simulator to decide between exploring related videos or searching for new queries based on historical preferences.}
\label{tbl:prompt-video-retrieve}
\end{table}

\begin{table}[h]
\centering
\small
\begin{tabular}{p{\linewidth}}
\toprule
\textbf{Prompt for Video Selection} \\
\midrule
You are a video quality rater, responsible for selecting the most relevant and least relevant videos based on your preferences. \\[1em]

Previously, you have watched the following videos: \\
\{history\} \\[1em]

So far, you have defined your preferences as: \\
\{preference\} \\[1em]

You now want to watch a video related to \{query\}. \\
From the list of candidate videos, choose the most wanted video (the one that best matches your preferences and the query) and the least wanted video (the one that least matches your preferences and the query). \\
Explain why each video is the most or least relevant to your preferences and the query. \\[1em]

Index starts from 1. (If you want to select the first video, you should write 1, not 0.) \\
If there is no appropriate candidate for the most or least wanted video, you should write [None]. \\[1em]

Candidate Videos: \\
\{candidate\} \\[1em]

Answer Format: Fill [] with your response. Do not return anything else. \\[1em]

Most Wanted: [video number] \\
Explanation: [Why this video best matches your preferences and the query] \\[1em]

Least Wanted: [video number] \\
Explanation: [Why this video least matches your preferences and the query] \\
\bottomrule
\end{tabular}%
\caption{Instructions for an LLM-based user simulator to select the most and least wanted videos based on preferences and a query.}
\label{tbl:prompt-video-selec}
\end{table}

\begin{table}[h]
\centering
\small
\resizebox{0.99\linewidth}{!}{%
\begin{tabular}{p{\linewidth}}
\toprule
\textbf{Prompt for \textit{Watching} a Video} \\
\midrule
You are a YouTube viewer with your preferences, and you should create a video summary based on how well it aligns with your personal preferences. \\[1em]

Context: Your latest updated preferences are as follows: \\
\{preference\} \\[1em]

Now, you are watching a new video, presented as a series of (frame description, transcript) pairs. \\[1em]

Video: \\
\{video\} \\[1em]

Write your summary of the video, followed by your personal opinion of the video. \\
The summary and personal opinion should be 2 sentences each. \\
For personal opinion, you may refer to 1 or 2 preferences that are related to reviewing the video. \\
But make sure your opinion should be mainly based on the video content, not just your preferences. You may like or dislike the video. \\
Return only one paragraph for the answer. \\
\bottomrule
\end{tabular}%
}
\caption{Instructions for an LLM-based user simulator to summarize and review a video based on its content and user preferences.}
\label{tbl:prompt-video-watch}
\end{table}

\begin{table}[h]
\centering
\small
\resizebox{0.99\linewidth}{!}{%
\begin{tabular}{p{\linewidth}}
\toprule
\textbf{Prompt for Preference Update} \\
\midrule
\begin{minipage}[t]{\linewidth}
You are a preference analyzer, responsible for refining user preferences based on user-written reviews and reasons for the most and least wanted videos.

\vspace{1em}

Your preferences have previously been defined as: \\
\{preference\}

\vspace{1em}

Next, your previous reviews on videos you watched are as follows: \\
\{reviews\}

\vspace{1em}

Next, you selected the following video as the most wanted video based on your preferences: \\
\{selected video\} \\
Reason: \\
\{selected reason\}

\vspace{1em}

Next, you selected the following video as the least wanted video based on your preferences: \\
\{least wanted video\} \\
Reason: \\
\{least wanted reason\}

\vspace{1em}

Based on the previous reviews and reasoning for the most and least wanted videos, re-define your preferences and dis-preferences in bullet points.

\vspace{1em}

Let’s break it down step by step:

\begin{enumerate}
    \item \textbf{ADDITION}: Add new details that were introduced in the video and are not in your previous preferences.
    \item \textbf{REFINEMENT}: Refine or adjust your preferences with specific terms where necessary to encompass both your original preferences and new insights from the video. Unless the content itself needs to change, reuse the exact words from the existing preferences.
    \item \textbf{REMOVAL}: Remove any details from your preferences where the newly watched video and your original preferences do not align.
\end{enumerate}

\end{minipage}
\\
\bottomrule
\end{tabular}%
}
\caption{Instructions for an LLM-based user simulator to update user preferences based on previous video reviews and reasoning.}
\label{tbl:prompt-preference-update}
\end{table}

\begin{table}[h]
\centering
\small
\resizebox{0.99\linewidth}{!}{%
\begin{tabular}{p{\linewidth}}
\toprule
\textbf{Prompt for Scoring Video Clips Based on Viewer Preferences} \\
\midrule
You are a viewer with specific content preferences. Evaluate multiple video clips and provide a score from 1 to 10 based on their appeal to you. \\[1em]

{Preference Profile:} \\
\{preference\_profile\} \\[1em]

For each clip below, determine how appealing it would be to you. Consider engagement, pacing, and overall impact. Provide a score and a short justification for each clip. \\[1em]

{Clips to Evaluate:} \\
\{clip info\}
\\[1em]

\textbf{Output Format:} \\
A list of responses for each clip, using this format: \\[0.5em]
- Clip ID: clip\_0, Score: 8, Justification: "Interesting content and engaging pacing." \\
- Clip ID: clip\_1, Score: 5, Justification: "Too slow and not aligned with my interests." \\
\bottomrule
\end{tabular}%
}
\caption{Instructions for an LLM-based user simulator to rate the appeal of video clips based on personal preferences and provide justifications.}
\label{tbl:prompt-score}
\end{table}

\subsection{Details of Human Verification}
\label{apdx:human_verf}
To assess the reliability and plausibility of the watch history generated by our LLM-based user simulator, we conducted human evaluations on two key components of the simulation framework: (1) query generation and (2) video selection. These evaluations were performed using Amazon Mechanical Turk (MTurk), with three independent annotators assigned to each task.

Annotators were provided with the exact same input as the LLM-based user simulator—namely, the set of previously watched videos and the user’s long-term preference at the time of decision. For video selection, annotators were also given the same video candidate pool, including metadata such as titles, thumbnails, and descriptions (as detailed in Appendix~\ref{apdx:youtube_crawling}).

\smallsection{Query Generation}
For each query generated by the simulator, annotators were asked if this is a reasonable next search query given the user’s watch, search history, and preference.
Each response was labeled as either reasonable or not reasonable. 
The agreement rate refers to the proportion of queries that were rated reasonable by the majority (i.e., at least two out of three annotators), which amounted to 97.56\%.
We also report inter-annotator agreement, computed as Fleiss’ $\kappa$, which was 0.85, indicating substantial agreement among annotators and reinforcing the consistency of the task design and the reliability of the simulator’s outputs.

\smallsection{Video Selection}
Annotators were asked to select the most preferred video from a set of candidates based on the provided preference information. The simulator’s choice was then compared to the majority choice of the annotators. 
If at least two annotators selected the same video as the simulator, the decision was considered a match.
The agreement rate between the simulator and human annotators was 68.42\%, showing that the simulator aligned with human choices in the majority of cases.
This demonstrates its ability to mimic realistic user behavior when making preference-based decisions.

These results validate the plausibility of the generated watch histories and support the reliability of the simulation framework used in constructing \dataset. 
While simulated behavior cannot perfectly replicate real user actions, our evaluations suggest that the LLM-based user simulator closely approximates human decision-making and offers a scalable foundation for studying personalized video summarization.

\subsection{Details on Analyses}

\smallsection{Inter-history video diversity}
To generate an embedding for each watch history, we first extract visual features from each video using CLIP-ViT/B-32. 
These features are then averaged per video to capture its overall content, resulting in a representative feature for each video. 
Finally, we average the features across all videos in watch history, yielding a single embedding for the entire history. 
This embedding serves as a compact representation of the user’s viewing patterns and preferences, capturing both individual video content and the broader interests reflected in the sequence of watched videos.

\smallsection{Saliency Score Distribution}
Figure~\ref{fig:score_distb} shows the distribution of saliency scores using kernel density estimation (KDE). The mean saliency score (left) is computed as the average of the saliency scores for all segments in a video, given by the formula: $\text{mean} = \frac{1}{n} \sum_{k=1}^{n} y_k$ where $y_k$ is the saliency score for segment $k$, and $n$ is the total number of segments in the video. The mean provides a measure of the average relevance of the segments within a video. In our dataset, most videos have a mean score between 4 and 6, indicating that the majority of videos are considered moderately important overall.

To evaluate how much the saliency scores fluctuate across segments, we compute the standard deviation, which measures the spread of the scores around the mean. The standard deviation is calculated as: $\text{std} = \sqrt{\frac{1}{n} \sum_{k=1}^{n} (y_k - \text{mean})^2}$
where $y_k$ represents the individual saliency score for segment $k$, and $\text{mean}$ is the mean saliency score for the video. The standard deviation quantifies the variability in the importance of the segments. Standard deviations typically range from 1.5 to 2 in our dataset, suggesting moderate variability in how segments are scored. Higher standard deviations (greater than 3) indicate greater fluctuations in segment importance, which may result from dynamic visual changes, such as scene transitions or shifts in the video’s content, where some segments may be much more relevant than others.

\subsection{Long-term Preference Modeling}
Table~\ref{tbl:long-term-pref} illustrates how the LLM-based user simulator refines its long-term preferences as it iteratively interacts with more videos.
Initially, the preferences are broad and general, but as the simulator processes more content, they become increasingly specific and nuanced. 
Through repeated engagement with the content, the simulator develops a detailed preference model that captures both high-level interests and subtle distinctions, emphasizing the importance of iterative interactions in accurately modeling complex user preferences.

\begin{table}[t]
\centering
\begin{tabular}{p{0.48\textwidth} p{0.48\textwidth}}
\toprule
\multicolumn{2}{l}{{Watch History (0 Videos)}} \\
\midrule
\multicolumn{2}{l}{
\hspace{1em}• I like {recent news} on {Job}, especially about {Lawyer}. } \\
\midrule
\multicolumn{2}{l}{{Watch History (3 Videos)}} \\
\midrule
{[ Likes ]} & {[ Dis-likes ]} \\
\hspace{1em}• {Informative insights} with {dynamic} and {visually appealing presentations}. 
& • Lack of {engaging visuals} and {narrative structure}. \\
\hspace{1em}• Interest in {latest trends} and {technology} in the {legal profession}, 
especially related to {law firm management}, including {generative AI}. 
& • Does not directly address {current trends} affecting the {legal profession}. \\
\midrule

\multicolumn{2}{l}{{Watch History (10 Videos)}} \\
\midrule
{[ Likes ]} & {[ Dis-likes ]} \\
\hspace{1em}• Combines {informative insights} with {dynamic}, {visually appealing presentations}. 
& • Lacks {engaging visuals} and {narrative structure}. \\
\hspace{1em}• Interest in {technology} and {trends} in the {legal profession}, including 
{generative AI}, {LawTech}, and {RegTech}. 
& • Omits discussion of {technological trends}, including those influenced by the {pandemic}. \\
\hspace{1em}• Prefers a {cohesive flow} and {engaging narrative} exploring {risks} and 
{ethical challenges} of {AI}. 
& • Too focused on {patent examination} without broader {technological context}. \\
\hspace{1em}• Values insights on how {law firms} use {AI} to improve {services} and move beyond 
{billable hour models}. 
& • Emphasizes {marketing/scaling} law firms without connecting to {tech impact}. \\
\hspace{1em}• Enjoys {practical applications} of {AI} and their role in advancing {legal workflows} 
and {access to justice}. & \\
\hspace{1em}• Likes content highlighting how {technology} supports {underserved populations} and enables 
{judicial reform}. & \\
\hspace{1em}• Interested in the {ethical implications} of {AI} and its effect on {trust} in 
the legal system. & \\
\bottomrule
\end{tabular}
\caption{Long-term user preferences for video content related to legal professions and technological advancements, organized by watch history length.}
\label{tbl:long-term-pref}
\end{table}

\section{More Details on Experiments}

\subsection{Baseline Details}
\label{apdx:baselines}
\textbf{SL-module} replaces conventional pair-based learning in video highlight detection with a set-based approach. Instead of comparing segment pairs, it evaluates a set of video segments to predict highlight scores by modeling inter-dependencies among segments within the same video.
A fixed visual feature extractor processes each segment, followed by a transformer encoder (without positional encoding) to capture contextual relationships. The transformer output is passed to a scoring model that outputs highlight scores. The model is trained by minimizing the KL divergence between predicted and ground-truth highlight score distributions over the set.

\textbf{Moment-DETR} projects video and query features into a shared embedding space, concatenates them, and processes the result with a transformer encoder using positional encodings. A linear layer predicts saliency scores from the encoder output. A transformer decoder, initialized with moment queries, predicts temporal moments; its output feeds into a three-layer FFN for normalized coordinates and a softmax classifier for moment-level scores.

\textbf{UMT} starts by processing visual and audio features with separate transformer encoders. Then, a bottleneck transformer fuses these features into multi-modal representations. If a text query is provided, it is used to generate temporally-aligned, clip-specific moment queries via attention between the text and the multi-modal features. The generated queries are then decoded to obtain joint representations for both tasks. Finally, the model produces clip-level saliency scores for highlight detection and moment boundaries (center, window, and offset) for moment retrieval.

\textbf{VSL} suummarizes videos based on user-preferred genres using a similarity-based approach. It generates scene-level textual summaries from visual captions and transcripts, then computes similarity between these summaries and genre embeddings derived from text prompts.

\subsection{Evaluation Metrics}
\label{apdx:eval_metrics}

To comprehensively assess our model’s performance across tasks, we utilize several widely accepted evaluation metrics. These metrics are chosen based on the nature of each task—whether it involves ranking video segments, retrieving specific moments, or predicting saliency scores. Below, we explain each metric and its purpose in detail.

For highlight detection, the goal is to rank video segments based on how salient or important they are. To evaluate how well our model performs this ranking, two key metrics are usually used:
\begin{itemize}
    \item Mean Average Precision (mAP): mAP measures the quality of ranked results. It checks whether the most relevant (i.e., salient) segments appear near the top of the list. For each relevant segment, it calculates the precision (i.e., how many of the top-ranked results are correct), and then averages this across all relevant segments. Finally, the average is taken over all test samples. Higher mAP values indicate better performance, meaning the model ranks relevant segments closer to the top more consistently.
    \item Hit@1: This metric simply checks whether the top-ranked video segment is actually one of the ground truth salient segments. It’s a straightforward way to see if the model gets the very best answer right. 
\end{itemize}

To decide which segments are “\textit{salient},” we use saliency score thresholds. 
These scores are given on a scale from 1 to 10. 
Following the methodology of \citet{liu2015multi}, we treat segments with scores $\ge7$ and $\ge9$ as salient (i.e., ground truth highlights). 
In \citep{liu2015multi}, the threshold was 4 out of 5, which corresponds to a similar percentile cut-off.

In Moment Retrieval, the task is to retrieve a specific time segment in the video that corresponds to a given query (e.g., “when the player scores a goal”). Here, we want to know whether the model correctly identifies the right moment in time.
\begin{itemize}
    \item Recall@1: This metric evaluates whether the top segment predicted by the model has sufficient overlap with the ground truth moment. It focuses on the single top-ranked result. If this result matches the correct segment well enough, it’s counted as a success.
\end{itemize}

To define what counts as a good match, we use Intersection over Union (IoU), a standard metric in temporal and spatial localization tasks. IoU compares how much the predicted time span overlaps with the ground truth. It is calculated as the length of the overlap divided by the length of the union of both time spans.
If IoU $\ge 0.5$: The prediction is considered correct if at least 50\% of the predicted and ground truth segments overlap. If IoU $\ge0.7$, the match must be even more precise, with at least 70\% overlap.
Higher Recall@1 values mean that the model is retrieving relevant moments more accurately.

\subsection{Ablation Studies}
\begin{wrapfigure}{r}{0.35\textwidth}
    \centering
    \vspace{-0.5cm}
    \includegraphics[width=0.95\linewidth]{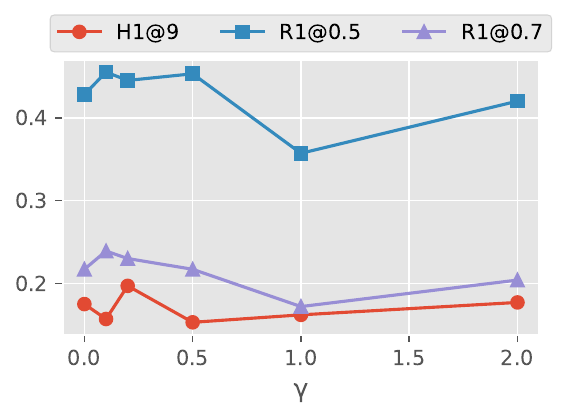}
    \vspace{-0.4cm}
    \caption{Performance of varying $\gamma$.}
    \vspace{-0.7cm}
    \label{fig:gamma_ablation}
\end{wrapfigure}
While we initially set $\gamma$ = 1 following prior works~\citep{lei2021detecting,moon2023query}, we conducted an ablation study to assess its impact on \proposed’s performance. 
We observed that smaller margins ($\gamma$= 0.1–0.2) consistently yield better results, as they enable finer-grained preference modeling. In contrast, larger margins tend to encourage overconfident separation between segments, which reduces generalization. We will include these ablation results in the final version to specify the impact of $\gamma$.

\subsection{Case Study}
\label{apdx:case_studies}
\begin{figure*}[h]
    \centering    
    \includegraphics[width=\linewidth]{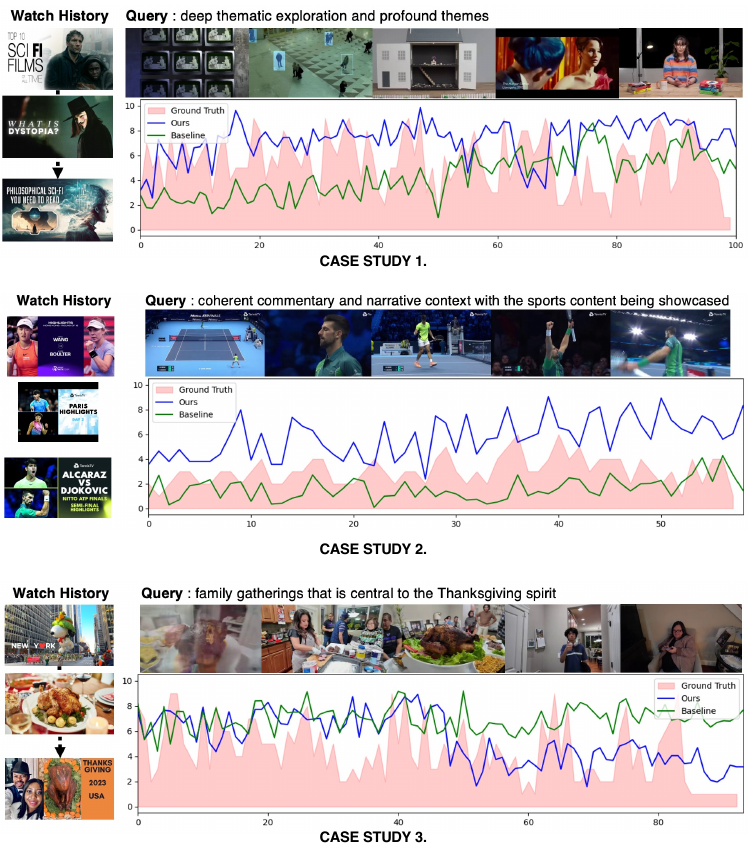}
    \caption{Qualitative Case Studies: “Ours” refers to \proposed, and “Baselines” refers to Moment-DETR.}
    \label{fig:case_studies}
\end{figure*}

Figure~\ref{fig:case_studies} presents qualitative case studies across various video topics (e.g., movies, sports, holidays), comparing the predicted scores of \proposed and the baseline (Moment-DETR).

\end{document}